\documentclass[11pt]{article}

\usepackage[preprint]{acl}
\usepackage{times}
\usepackage{latexsym}
\usepackage{amsmath}
\usepackage{amssymb}
\usepackage{kotex}
\usepackage{subcaption}
\usepackage{subcaption}
\usepackage{makecell}
\usepackage{enumitem}

\usepackage[T1]{fontenc}

\usepackage[utf8]{inputenc}

\usepackage{microtype}

\usepackage{inconsolata}

\usepackage{graphicx}

\newcommand{\defeq}{\overset{\text{\tiny def}}{=}}

\usepackage{booktabs}
\usepackage[table]{xcolor}
\usepackage{multirow}
\usepackage{array}
\usepackage{float}

\definecolor{bestblue}{RGB}{155,194,230}
\definecolor{secondblue}{RGB}{221,235,247}

\title{Correctness-Dependent Asymmetric Target Distributions \\
for Self-Distillation Policy Optimization}

\author{
  \textbf{Suwan Yoon}\textsuperscript{1*},
  \textbf{Seongryong Jung}\textsuperscript{2*},
  \textbf{Changhee Lee}\textsuperscript{1$\dagger$}
  \\[0.5em]
  \textsuperscript{1}Korea University, Seoul, Korea,
  \textsuperscript{2}Chung-Ang University, Seoul, Korea
  \\
  \texttt{\{suwanyoon, changheelee\}@korea.ac.kr}
  \\
  \texttt{jungsr1116@cau.ac.kr}
}

\usepackage[most]{tcolorbox}

\newtcolorbox{propositionbox}{
    colback=gray!8,
    colframe=gray!35,
    boxrule=0.5pt,
    arc=2mm,
    left=2mm,
    right=2mm,
    top=1.5mm,
    bottom=1.5mm,
    breakable,
    before skip=6pt,
    after skip=6pt
}

\usepackage{mdframed}

\mdfdefinestyle{mathblockframe}{
  linecolor=gray!40,
  linewidth=0.5pt,
  backgroundcolor=gray!8,
  roundcorner=3pt,
  innertopmargin=5pt,
  innerbottommargin=5pt,
  innerleftmargin=3pt,
  innerrightmargin=3pt,
  skipabove=3pt,
  skipbelow=3pt,
}

\newmdtheoremenv[style=mathblockframe]{theorem}{Theorem}
\newmdtheoremenv[style=mathblockframe]{proposition}[theorem]{Proposition}
\newmdtheoremenv[style=mathblockframe]{corollary}[theorem]{Corollary}
\newmdtheoremenv[style=mathblockframe]{observation}[theorem]{Observation}

\begin{document}
\maketitle

\begingroup
\renewcommand{\thefootnote}{\fnsymbol{footnote}}
\footnotetext[1]{These authors contributed equally.
Their author order was determined by a coin toss.}
\footnotetext[2]{Corresponding author.}
\endgroup

\begin{abstract}
Self-distillation policy optimization provides dense token-level supervision for LLM post-training through feedback-conditioned self-teachers. However, directly matching the self-teacher can disrupt successful behaviors on correct rollouts while providing insufficient correction on incorrect ones. We propose \textbf{Asymmetric Policy Optimization via Self-Distillation (APSD)}, which constructs correctness-dependent targets by mixing student and self-teacher distributions: interpolation preserves successful behaviors on correct rollouts, while extrapolation beyond the self-teacher strengthens correction on incorrect ones. We analyze APSD through divergence decomposition, reverse-KL barycentric geometry, and gradient analysis, highlighting the importance of gradient flow through student-dependent targets. Experiments under reinforcement learning with verifiable rewards and rich feedback show that APSD achieves strong overall performance against reinforcement-learning and self-distillation baselines and matches SDPO performance with 2.6$\times$ fewer training steps on average across Qwen3 model scales. Code is available \textcolor{blue}{\underline{\href{https://anonymous.4open.science/r/APSD-0/}{here}}}.

\end{abstract}

\section{Introduction}

\begin{figure}[t]
    \centering
    \makebox[\columnwidth][c]{%
        \includegraphics[width=1.0\columnwidth]{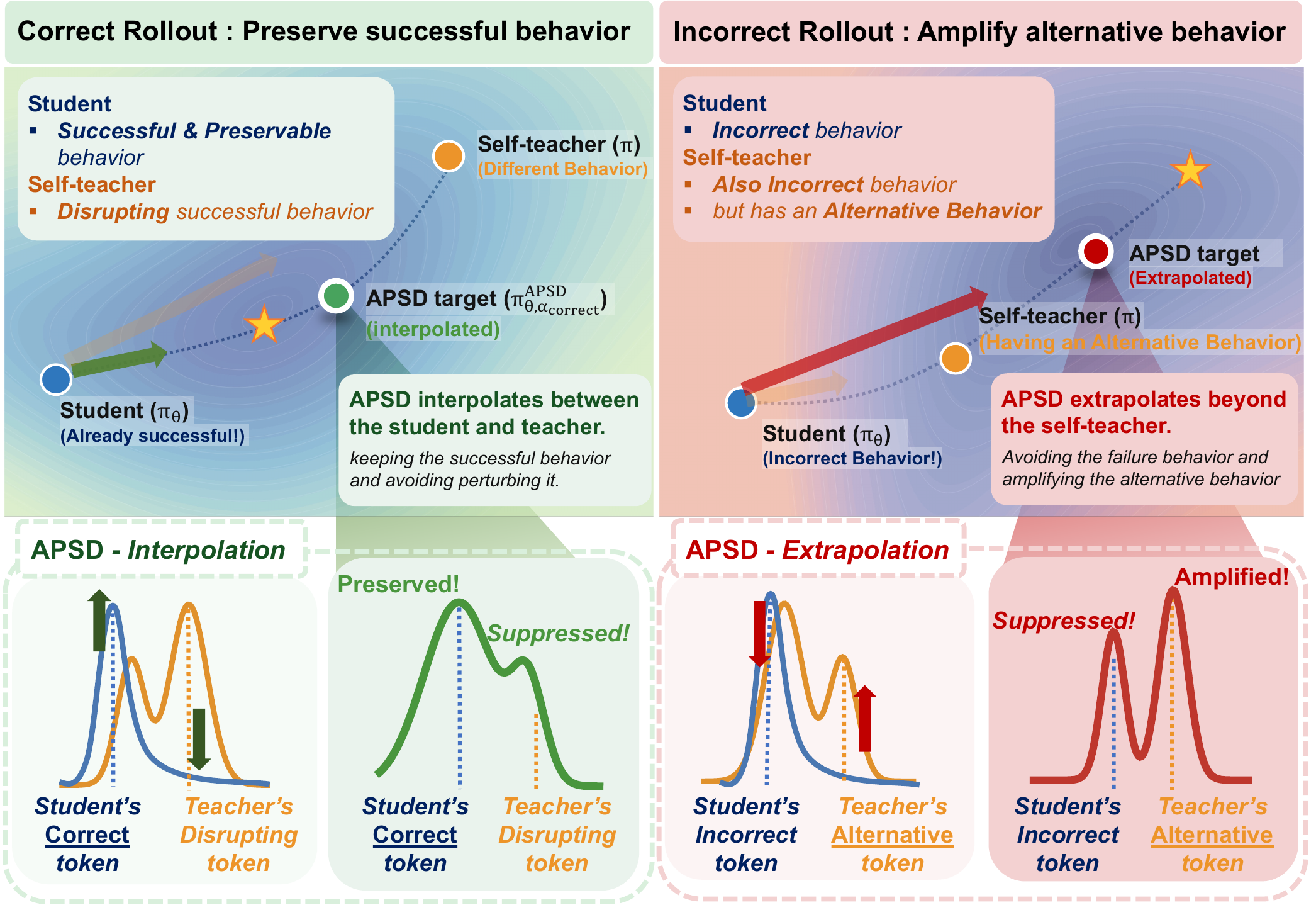}%
    }
    \vspace{-1.5em}
    \caption{\textbf{Correctness-dependent target construction in APSD.}
    Directly matching the self-teacher can disrupt successful behaviors in correct rollouts while providing insufficient correction for incorrect ones.
    APSD addresses this asymmetry by adapting the distillation target according to rollout correctness. }
    \vspace{-1.5em}
    \label{fig:intro}
\end{figure}

Reinforcement learning with verifiable rewards (RLVR) has emerged as a prominent post-training paradigm for enhancing the reasoning and problem-solving capabilities of LLMs~\citep{shao2024deepseekmathpushinglimitsmathematical, Guo_2025, kimiteam2025kimik15scalingreinforcement}. However, RLVR typically relies on scalar outcome-level rewards, providing limited, coarse-grained guidance for assigning credit to individual tokens in a generated response. To address this limitation, recent self-distillation approaches leverage privileged information to provide fine-grained, token-level supervision beyond scalar rewards~\citep{ye2026onpolicycontextdistillationlanguage,shenfeld2026selfdistillation,zhao2026selfdistilled,hubotter2026reinforcement,song2026expanding}. This paradigm naturally extends to reinforcement learning with rich feedback (RLRF), where textual environment feedback provides additional learning signals beyond outcome rewards. These approaches construct a self-teacher by conditioning the same underlying policy on privileged information and distill its token-level distributions into the student policy.


Building on this paradigm, recent studies have incorporated rollout correctness into self-distillation to regulate teacher-guided supervision for successful and unsuccessful trajectories~\citep{yang2026selfdistilledrlvr,li2026unifying,kim2026rebellious}. Meanwhile, research on knowledge distillation has explored alternative target distributions that combine teacher and student distributions to improve knowledge transfer~\citep{ko2024distillm,shing2025taid,shin2026amid,yang2026learningteachergeneralizedonpolicy}. Despite these advances, how rollout correctness should determine the construction of distillation targets relative to the student and self-teacher distributions remains largely underexplored.



To address this blind spot, we analyze the token-level discrepancies between the student and its feedback-conditioned self-teacher across correct and incorrect rollouts. Our analysis reveals contrasting patterns: for correct rollouts, the self-teacher can substantially deviate from the student's already successful token preferences, potentially disrupting desirable behaviors. For incorrect rollouts, however, the self-teacher can remain close to the student distribution, providing insufficient separation to correct failed behaviors effectively. These observations reveal a mismatch between the teacher–student discrepancy and the degree of correction required by the student, as conceptually illustrated in Figure~\ref{fig:intro}. This motivates \textit{a correctness-aware strategy} that preserves successful student behaviors while providing stronger and precise corrective guidance for unsuccessful ones.

To address this correctness-dependent asymmetry, we propose \textbf{Asymmetric Policy Optimization via Self-Distillation (APSD)}, which constructs asymmetric distillation targets according to rollout correctness. Specifically, APSD constructs a target distribution by reweighting the student distribution using the teacher-to-student probability ratio, with a correctness-dependent coefficient controlling the relative influence of teacher- and student-preferred tokens. As illustrated in Figure~\ref{fig:intro}, APSD interpolates between the student and self-teacher distributions for correct rollouts, incorporating teacher feedback while preserving the student's successful token preferences. For incorrect rollouts, APSD instead extrapolates beyond the self-teacher distribution, shifting the target further away from the student and toward alternatives favored by the teacher.

Our theoretical analyses show that the APSD objective decomposes into a \textit{reverse-KL (RKL) term and a counterbalancing Rényi-divergence term}, while the asymmetric coefficient controls \textit{target interpolation and extrapolation} along a generalized RKL barycentric path. Crucially, we show that \textit{preserving gradient flow through the student-dependent target} retains an additional coupling term, whereas fully detaching the target reduces each token-level gradient to a scaled SDPO gradient. Experiments across RLVR and RLRF demonstrate that APSD consistently outperforms state-of-the-art reinforcement-learning and self-distillation baselines while improving training efficiency. Further controlled sampling experiments and ablation studies support the effectiveness of correctness-dependent target construction and highlight the importance of preserving gradient flow.


\begin{figure*}[t]
    \centering
    \includegraphics[width=0.80\textwidth]
    {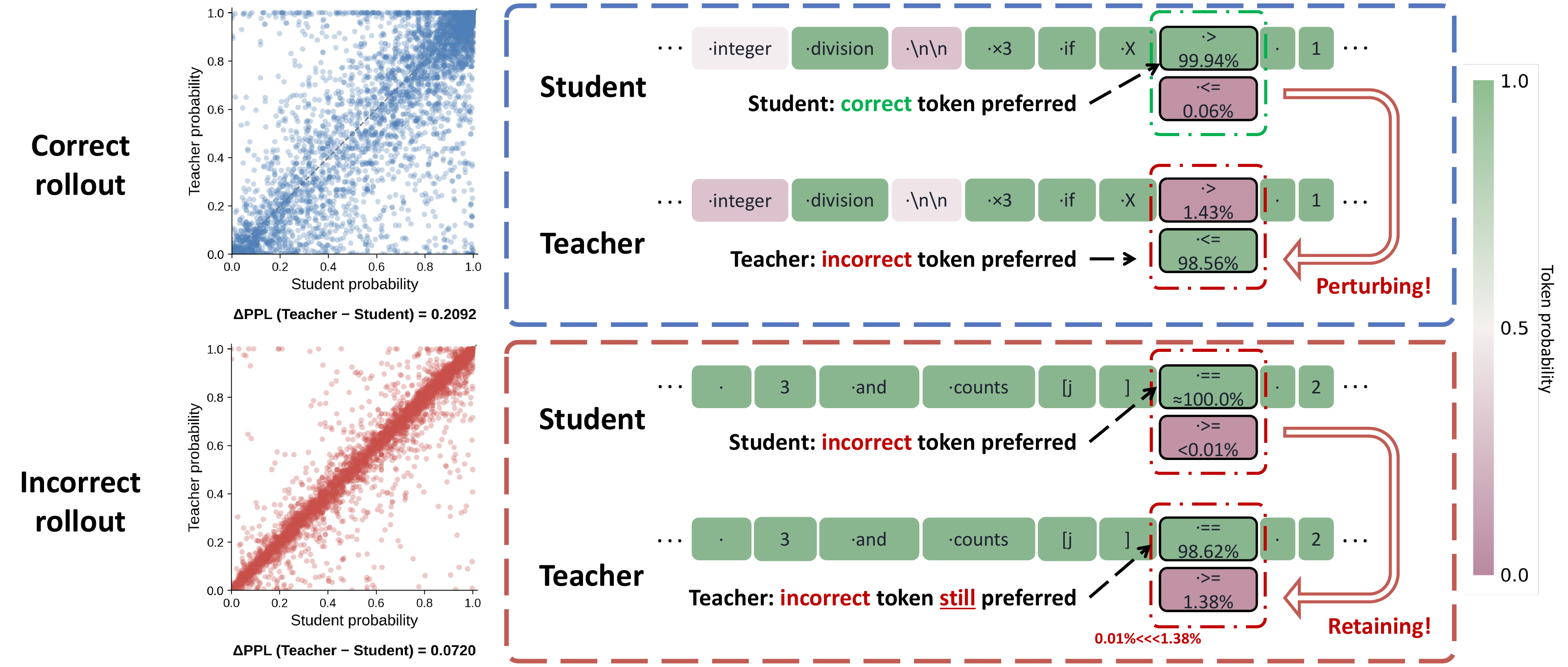}
    \vspace{-0.5em}
    \caption{
    Teacher--student discrepancy in SDPO on LiveCodeBench v6.
    The left panels show token-probability discrepancies for correct and incorrect
    rollouts, while the right panels provide representative examples of perturbing
    a correct choice and retaining an incorrect choice.
    }
    \vspace{-1.5em}
    \label{fig:motivation}
\end{figure*}

\section{Related Work}

\paragraph{Self-Distillation for LLM Post-Training.}

Self-distillation has emerged as an effective approach to LLM post-training, where privileged self-teachers constructed from the same underlying policy provide fine-grained supervision using additional context or environment feedback~\citep{ye2026onpolicycontextdistillationlanguage,shenfeld2026selfdistillation,zhao2026selfdistilled,hubotter2026reinforcement,song2026expanding}.
In particular, SDPO~\citep{hubotter2026reinforcement} conditions the self-teacher on textual environment feedback, converting rich feedback into token-level supervision beyond scalar rewards.
Subsequent work incorporates rollout rewards or correctness to regulate self-teacher supervision, highlighting the importance of distinguishing successful and failed rollouts~\citep{yang2026selfdistilledrlvr,li2026unifying,kim2026rebellious}.

\paragraph{Distribution Mixing for Distillation.}

Standard knowledge distillation transfers knowledge from a teacher to a student by using the teacher's output distribution as supervision~\citep{hinton2015distillingknowledgeneuralnetwork, gu2024minillm, jung-etal-2025-todi}.
Recent work has extended this paradigm by constructing alternative distillation targets from the student and teacher distributions to improve distillation effectiveness~\citep{ko2024distillm,shing2025taid,shin2026amid,yang2026learningteachergeneralizedonpolicy}.
Similar ideas have been explored in self-distillation for LLM post-training by combining the student or reference policy with a privileged self-teacher~\citep{xu2026betaopsdderivingpolicyoptimization,li2026demopsddisagreementmodulatedpolicyselfdistillation}.


\paragraph{Our Work.}

While prior work has explored correctness-aware supervision strategies and teacher--student distribution mixing, how rollout correctness should govern the direction and geometry of distillation target construction remains underexplored. APSD bridges this gap by using rollout correctness to determine whether the target interpolates between or extrapolates beyond the student and self-teacher distributions. In addition, APSD preserves gradient flow through its student-dependent target, preventing the APSD gradient from collapsing onto the SDPO update direction.

\section{Correctness-Dependency in SDPO} \label{sec:problem_formulation}
\paragraph{Notations.}
Let $x$ denote the input prompt and $\pi_\theta$ denote the current student policy. Given $x$, we sample a group of $G$ rollouts from $\pi_\theta$, denoted by $\{y^i\}_{i=1}^{G}$. Each rollout is a token sequence $y^i=(y_1^i,\ldots,y_T^i)$, where $y_t^i$ is the $t$-th token and $y_{<t}^i$ denotes the preceding prefix. Let $f^i$ denote the textual environment feedback associated with rollout $y^i$ for input $x$ (e.g., a successful rollout from the same rollout group and a code execution error log). We use $\mathcal{V}$ to denote the vocabulary.

\paragraph{SDPO.} For each rollout $y^i$, the student distribution at decoding step $t$ is denoted by $\pi_\theta(y_t^i \mid x, y_{<t}^i)$. SDPO constructs a corresponding self-teacher distribution by conditioning on the feedback $f^i$ as $\pi(y_t^i \mid x, y_{<t}^i) \defeq \texttt{sg}[\pi_\theta(y_t^i \mid x, f^i, y_{<t}^i)]$, where $\texttt{sg}[\cdot]$ 
denotes the stop-gradient operator. 
For notational simplicity, we henceforth omit the conditioning variables and denote the student and self-teacher
distributions by $\pi_\theta$ and $\pi$, respectively.

Under this formulation, SDPO directly adopts the self-teacher distribution as the distillation target, i.e., $\pi_{\mathrm{target}}=\pi$, and minimizes the reverse KL (RKL) divergence~\citep{gu2024minillm} between the student and target distributions:
\begin{equation} \label{eq:spdo}
\mathcal{L}_{\mathrm{SDPO}}(\theta) = \sum_t D_{\mathrm{KL}} \left( \pi_\theta \middle\| \pi_{\mathrm{target}} \right),
\end{equation}
where $D_{\text{KL}}(p\| q)=\sum_ip(i){p(i)\over q(i)}. $
By conditioning the self-teacher on textual feedback $f^i$, SDPO converts rich environmental feedback into fine-grained, token-level supervision. This provides a more informative learning signal than the trajectory-level reward, and has proven effective in settings with verifiable rewards and rich feedback.

\paragraph{Challenges.}
Although SDPO provides dense supervision by distilling toward a feedback-conditioned self-teacher, using the self-teacher distribution as a uniform target across rollouts may overlook an important distinction that the desired policy update should depend on rollout correctness. As illustrated in Figure~\ref{fig:motivation}, we observe two contrasting failure cases: 
\begin{itemize}[leftmargin=10pt,noitemsep,topsep=0pt] \vspace{0.5mm}
    \item \textbf{Disrupting successful behaviors.} For correct rollouts, the self-teacher may deviate from the student's successful token preferences, potentially disrupting desirable behavior.
    \vspace{1mm}
    \item \textbf{Insufficient correction of failed behaviors.} 
    For incorrect rollouts, the self-teacher may remain close to the student's erroneous preferences, providing insufficient guidance toward better alternatives.
    \vspace{1mm}
\end{itemize}

These observations motivate an asymmetric adaptation of the distillation target based on rollout correctness. For correct rollouts, the target should remain closer to the student distribution, preserving successful behaviors while incorporating feedback from the self-teacher. For incorrect rollouts, the target should move beyond the self-teacher distribution away from the student, placing greater emphasis on alternatives relatively favored by the self-teacher. This interpolation--extrapolation mechanism allows the target to preserve successful behaviors while providing stronger corrective guidance for unsuccessful ones.\footnote{
We further support this correctness-dependent asymmetry through
a controlled continuation experiment in
Section~\ref{sub:sampling}.
See Appendix~\ref{app:teacher_student_analysis} for additional analyses of teacher--student discrepancy and
correctness-dependent APSD targets. 
}


\begin{figure*}[t]
    \centering
    \includegraphics[width=0.9\textwidth]{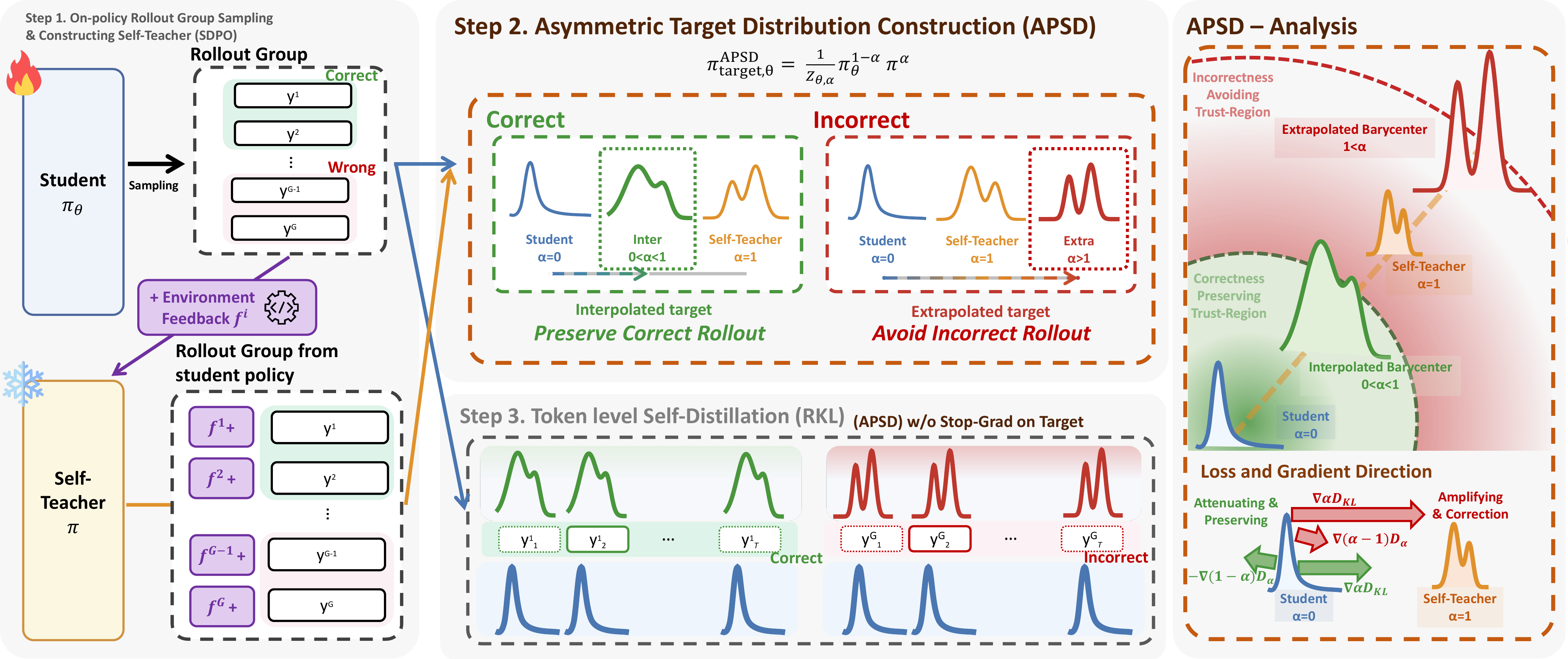} 
    \vspace{-0.6em}
\caption{Overview of APSD.
Step 1: Sample on-policy rollout groups and construct the self-teacher.
Step 2: Interpolate student--self-teacher targets for correct rollouts ($0<\alpha<1$) and extrapolate beyond the self-teacher for incorrect rollouts ($\alpha>1$).
Step 3: Token-level RKL self-distillation with gradients through the targets.
Analysis: The right panel shows target geometry and gradient directions.}
    \label{fig:APSD}
            \vspace{-1em}
\end{figure*}
\section{Method: APSD}
\label{method}
To operationalize this correctness-dependent asymmetry, we propose \textbf{Asymmetric Policy Optimization via Self-Distillation (APSD)}. In this section, we describe how APSD constructs asymmetric distillation targets to preserve the student's behavior on successful rollouts while providing stronger corrective guidance on failed ones. We then provide theoretical analyses to characterize this correctness-dependent behavior through divergence decomposition and gradient dynamics.

\subsection{Constructing Asymmetric Target Distribution}
\label{sub:apsd}

For each rollout $y^i$ in a group $\{y^i\}_{i=1}^{G}$ sampled from the current student policy $\pi_\theta$ given prompt $x$, we construct the self-distillation target $\pi_{\theta,\alpha}^{\mathrm{APSD}}$ by reweighting the student distribution according to the teacher-to-student probability ratio:
\begin{equation}
\label{eq:apsd_ratio}
\pi_{\theta,\alpha}^{\mathrm{APSD}}(v)
\propto
\pi_\theta(v)
\left(
\frac{\pi(v)}{\pi_\theta(v)}
\right)^{\alpha}~~\text{for}~v \in \mathcal{V}.
\end{equation}
Here, $\alpha \in \mathbb{R^+}$ is a correctness-dependent scalar that controls the asymmetry of the target.
The ratio $\pi(v)/\pi_\theta(v)$ serves as an importance weight, adjusting each token's probability according to its relative preference under the self-teacher and student. As $\alpha$ increases, the target places greater emphasis on tokens favored by the self-teacher relative to the student and suppresses those for which the self-teacher assigns lower relative probability.
 

Normalizing Eq.~\eqref{eq:apsd_ratio}, we express the APSD target as a generalized product-of-experts distribution over the student and self-teacher~\citep{qiu2025multimodal}:
\begin{equation} \label{eq:apsd_poe}
\pi_{\theta,\alpha}^{\mathrm{APSD}}(v) \!=\! \frac{1}{Z_{\theta,\alpha}} \pi_\theta(v)^{1-\alpha} \pi(v)^{\alpha}~~\text{for}~v \in \mathcal{V}, \!
\end{equation}
where $Z_{\theta,\alpha} = \sum_{u\in \mathcal{V}} \pi_\theta(u)^{1-\alpha} \pi(u)^{\alpha}$ is the normalizing constant. 
Here, $\alpha$ controls their relative contributions, inducing distinct target regimes:
\begin{itemize}[leftmargin=10pt,noitemsep,topsep=0pt] \vspace{0.5mm}
    \item \textbf{Interpolation} ($0 < \alpha < 1$): The target lies between the student and self-teacher distributions, preserving the student's preference while integrating feedback from the self-teacher.
    \item \textbf{Extrapolation} ($\alpha > 1$): The target extrapolates beyond the self-teacher distribution relative to the student, further amplifying tokens relatively favored by the self-teacher over the student. 
\end{itemize}
As reference cases, $\alpha=0$ recovers the student distribution, $\pi_{\theta,0}^{\mathrm{APSD}}=\pi_\theta$, while $\alpha=1$ recovers the self-teacher distribution,
$\pi_{\theta,1}^{\mathrm{APSD}}=\pi$.

To address the correctness-dependent discrepancies identified in Section~\ref{sec:problem_formulation}, we operationalize this asymmetric target by conditioning the choice of $\alpha$ according
to rollout correctness:
\begin{equation} \label{eq:apsd_target} 
\pi_{\mathrm{target},\theta}^\mathrm{APSD} =
\begin{cases}
\pi_{\theta,\alpha_{\mathrm{corr}}}^{\mathrm{APSD}}, & \text{if } y^i \text{ is correct},\\[1mm]
\pi_{\theta,\alpha_{\mathrm{incorr}}}^{\mathrm{APSD}}, & \text{otherwise}.
\end{cases}
\end{equation}
where $\alpha_{\mathrm{corr}}\in(0,1)$ and $\alpha_{\mathrm{incorr}}>1$.
This correctness-dependent target allows APSD to preserve successful student behaviors on correct rollouts while providing stronger corrective guidance on incorrect ones.
APSD then uses the SDPO distillation objective, replacing the self-teacher target with this correctness-dependent target:
\begin{equation} \label{eq:apsd_loss}
\mathcal{L}_{\mathrm{APSD}}(\theta) = \sum_t D_{\mathrm{KL}} \left( \pi_\theta \middle\| \pi_{\mathrm{target},\theta}^{\mathrm{APSD}} \right).
\end{equation}
Here, we maintain the dependence of $\pi_{\theta,\alpha}^{\mathrm{APSD}}$ on the current student $\pi_\theta$, allowing gradients to flow through the
student-dependent target. This gradient-preserving formulation is essential to APSD's asymmetric optimization behavior, which we analyze in Proposition~\ref{proposition3}.

\subsection{Theoretical Analysis}
\label{sec:theoretical-analysis}

In this subsection, we analyze APSD from three different perspectives: (i) decomposition of $\mathcal{L}_{\mathrm{APSD}}(\theta)$, (ii) geometry of $\pi_{\theta, \alpha}^\text{APSD}$, and (iii) dynamics of $\nabla_\theta \mathcal{L}_\text{APSD}(\theta)$. Together, these analyses characterize how $\alpha$ shapes the optimization objective, determines the target's position within the student--self-teacher RKL geometry, and influences gradient
propagation through the student-dependent target.

\paragraph{Perspective 1: Loss Decomposition.}\label{sub:Lenyi}
To analyze how $\alpha$ shapes the APSD objective in Eq.~\eqref{eq:apsd_loss}, we first decompose the loss into RKL and R\'enyi divergence terms. The following proposition establishes this decomposition, making explicit how $\alpha$ governs their relative contributions.

\begin{proposition}  \label{proposition1} 
\textbf{($\alpha$ Balances RKL and R\'enyi Divergences)}
For $\pi_{\theta,\alpha}^{\mathrm{APSD}}$ in Eq.~\eqref{eq:apsd_target}, the APSD loss in Eq.~\eqref{eq:apsd_loss} can be
decomposed as:
\begin{equation}
\label{eq:renyi_decomposition}
\!\!\mathcal{L}_{\mathrm{APSD}}
=
\alpha
D_{\mathrm{KL}}\!\left(\pi_\theta\middle\|\pi\right) - (1\!-\alpha) D_\alpha\!\left(\pi\middle\|\pi_\theta\right) \!, 
\end{equation}
where the latter term is the R\'enyi divergence of order $\alpha$~\citep{van2014renyi}:
\begin{equation} \nonumber
D_\alpha(\pi\|\pi_\theta)
=
\frac{1}{\alpha-1}
\log
\sum_{v\in\mathcal{V}}
\pi(v)
\left(
\frac{\pi(v)}{\pi_\theta(v)}
\right)^{\alpha-1}.
\end{equation}
Hence, $\alpha$ controls both the weight of the RKL term from SDPO in Eq.~\eqref{eq:spdo} and the sign and magnitude of the additional R\'enyi divergence term. See Appendix~\ref{app:renyi_derivation} for proof.
\end{proposition}

Proposition~\ref{proposition1} shows how the two terms contribute differently depending on rollout correctness. 
For correct rollouts, $\alpha_{\mathrm{corr}}\in(0,1)$, so the Rényi term has a negative coefficient in Eq.~\eqref{eq:renyi_decomposition}. In this regime, the factor
$\left(\pi(v)/\pi_\theta(v)\right)^{\alpha_{\mathrm{corr}}-1}$ places greater relative emphasis on tokens with low teacher-to-student probability ratios, i.e., tokens relatively favored by the student. The negatively weighted R\'enyi term thus counteracts the RKL pressure toward high-ratio regions, helping preserve successful student behavior from being overly displaced by teacher guidance.

In contrast, for incorrect rollouts, $\alpha_{\mathrm{incorr}}>1$, the Rényi term in Eq.~\eqref{eq:renyi_decomposition} has a positive coefficient. In this regime, the ratio-dependent factor increases with the teacher-to-student probability ratio, emphasizing regions favored by the self-teacher but under-covered by the student. Minimizing the positive R\'enyi term, together with the RKL term, encourages the student to shift probability mass toward these teacher-preferred alternatives, reducing their under-coverage in the student distribution.

\paragraph{Perspective 2: Target Position in RKL Geometry.}
To characterize the position of $\pi_{\theta,\alpha}^{\mathrm{APSD}}$ relative to the student and self-teacher, we examine its geometry using RKL divergence. The proposition shows that the APSD target is a generalized RKL barycenter connecting the two distributions.

\begin{proposition} \label{proposition2} 
\textbf{($\alpha$ Governs the Target Position in RKL Geometry)} 
For any $\alpha \in \mathbb{R}^+$, the APSD target in Eq.~\eqref{eq:apsd_target} satisfies
\begin{equation} \nonumber
\!\pi_{\theta,\alpha}^{\mathrm{APSD}}\! \!= \!\underset{r\in\Delta(\mathcal V)}{\text{\normalfont argmin}}~\mkern-2mu
(1-\alpha)D_{\mathrm{KL}}(r\|\pi_\theta)
+ \alpha D_{\mathrm{KL}}(r\|\pi)\mkern-2mu,
\end{equation}
where $\Delta(\mathcal V)$ is the probability simplex over the vocabulary $\mathcal{V}$. 
Thus, $\alpha$ parameterizes the target along the generalized RKL barycentric path connecting $\pi_\theta$ and $\pi$~\citep{qiu2025multimodal}.

\vspace{2mm}

\noindent
Furthermore, when $\pi_\theta \neq \pi$, the induced student-centered trust-region radius~\citep{schulman2015trust}, defined as $\epsilon_\theta(\alpha)=D_{\mathrm{KL}}(\pi_{\theta,\alpha}^{\mathrm{APSD}}|\pi_\theta)$, is strictly increasing in $\alpha$ for $\alpha > 0$. That is, increasing $\alpha$ pushes the target distribution monotonically farther from the current student along this barycentric path. See Appendix~\ref{app:barycenter} for proof.
\end{proposition}

Proposition~\ref{proposition2} provides a geometric interpretation of APSD's correctness-dependent target construction. For correct rollouts, $\alpha_{\mathrm{corr}}\in(0,1)$, both barycentric coefficients, $1-\alpha_{\mathrm{corr}}$ and $\alpha_{\mathrm{corr}}$, are positive. The target therefore interpolates between the student and self-teacher distributions, preserving the student's contribution while incorporating teacher feedback. Moreover, the monotonicity of $\epsilon_\theta(\alpha)$ ensures that the target remains closer to the student than the self-teacher does, i.e., $\epsilon_\theta(\alpha_{\mathrm{corr}}) < D_{\mathrm{KL}}(\pi\parallel\pi_\theta)$. This helps preserve successful reasoning behaviors while allowing the student to incorporate feedback without excessive deviation from its current policy.

\begin{table*}[t]
\small
\centering
\renewcommand{\arraystretch}{1}

\definecolor{BaseGray}{RGB}{244,244,244}

\definecolor{TopOne}{RGB}{240,220,120}
\definecolor{TopTwo}{RGB}{250,240,200}

\makebox[\linewidth][c]{%
\begin{tabular}{
    l
    *{6}{>{\centering\arraybackslash}m{1.45cm}}
}
\toprule

\textbf{Methods}
& \textbf{Chemistry}
& \textbf{Physics}
& \textbf{Biology}
& \textbf{Materials}
& \textbf{Tool use}
& \textbf{Average} \\

\midrule
\midrule

\rowcolor{gray!20}
\textbf{Olmo3-7B-Instruct}
& 24.02
& 38.20
& 16.13
& 33.91
& 38.88
& 30.23 \\

+ GRPO (off-policy)
& 59.70
& 62.81
& 54.75
& \cellcolor{TopTwo}\underline{78.52}
& 64.34
& 64.03 \\

+ GRPO
& 60.15
& 67.50
& 54.38
& 73.60
& 61.31
& 63.39 \\

+ SDPO
& \cellcolor{TopTwo}\underline{80.51}
& \cellcolor{TopOne}\textbf{72.34}
& 52.00
& 77.59
& 62.68
& 69.03 \\

+ SRPO
& 75.57
& \cellcolor{TopTwo}\underline{72.27}
& \cellcolor{TopOne}\textbf{64.12}
& 73.60
& 60.94
& \cellcolor{TopTwo}\underline{69.30} \\

+ RLSD
& 74.20
& 70.47
& 58.00
& 74.34
& \cellcolor{TopTwo}\underline{64.89}
& 68.38 \\

\textbf{+ APSD~(Ours)}
& \cellcolor{TopOne}\textbf{80.54}
& 70.94
& \cellcolor{TopTwo}\underline{60.00}
& \cellcolor{TopOne}\textbf{79.32}
& \cellcolor{TopOne}\textbf{65.44}
& \cellcolor{TopOne}\textbf{71.25} \\

\midrule
\midrule

\rowcolor{gray!30}
\textbf{Qwen3-8B}
& 41.46
& 60.55
& 32.88
& 59.64
& 58.18
& 50.54 \\

+ GRPO (off-policy)
& 77.02
& 74.69
& 57.38
& 77.92
& 66.82
& \cellcolor{TopTwo}\underline{70.77} \\

+ GRPO
& 65.18
& \cellcolor{TopTwo}\underline{75.55}
& 59.00
& 71.01
& 63.42
& 66.83 \\

+ SDPO
& 79.79
& 74.38
& 55.25
& 76.00
& \cellcolor{TopTwo}\underline{67.83}
& 70.65 \\

+ SRPO
& \cellcolor{TopTwo}\underline{81.31}
& 73.98
& 60.75
& 72.27
& 64.89
& 70.64 \\

+ RLSD
& 71.40
& 72.58
& \cellcolor{TopTwo}\underline{62.38}
& \cellcolor{TopOne}\textbf{78.79}
& 61.67
& 69.36 \\

\textbf{+ APSD~(Ours)}
& \cellcolor{TopOne}\textbf{82.05}
& \cellcolor{TopOne}\textbf{77.11}
& \cellcolor{TopOne}\textbf{71.63}
& \cellcolor{TopTwo}\underline{78.46}
& \cellcolor{TopOne}\textbf{69.12}
& \cellcolor{TopOne}\textbf{75.67} \\

\bottomrule
\end{tabular}%
}
\vspace{-0.5em}
\caption{
RLVR performance on five benchmarks. We report the highest Avg@16 accuracy (\%) achieved within a 5-hour
wall-clock training budget.
Average denotes the mean across all five benchmarks.
The best and second-best results within each base model are highlighted in
{\setlength{\fboxsep}{1pt}\colorbox{TopOne}{\textbf{dark yellow}}}
and
{\setlength{\fboxsep}{1pt}\colorbox{TopTwo}{\underline{light yellow}}},
respectively.
}
\label{tab:rlvr_table}
\vspace{-1.5em}
\end{table*}

In contrast, for incorrect rollouts, $\alpha_{\mathrm{incorr}}>1$, the student coefficient becomes negative, while the self-teacher
coefficient is greater than one. The target thus extrapolates beyond the self-teacher along the generalized RKL barycentric path, moving farther
from the student. Correspondingly, its student-centered KL divergence exceeds that of the self-teacher, i.e., $\epsilon_\theta(\alpha_{\mathrm{incorr}}) > D_{\mathrm{KL}}(\pi\parallel\pi_\theta)$. This allows a larger divergence from the current policy, enabling stronger corrective updates to move the student away from unsuccessful behaviors.

\paragraph{Perspective 3: Gradient Dynamics.}
\label{sub:SG}

Unlike prior methods that detach the entire target distribution \citep{shing2025taid,li2026demopsddisagreementmodulatedpolicyselfdistillation}, APSD preserves the dependence of $\pi_{\theta, \alpha}^{\mathrm{APSD}}$ on the student distribution. This distinction is crucial for allowing the asymmetric gradient dynamics induced by $\alpha$.


\begin{proposition} \label{proposition3}
\textbf{(Target Detachment Collapses to Scaled SDPO)}
With $\alpha$ and the self-teacher $\pi$ fixed during differentiation, applying a stop-gradient operator to the entire target $\pi_{\theta, \alpha}^{\mathrm{APSD}}$ provides
\begin{equation} \label{eq:SG}
\begin{aligned}
\nabla_\theta \mathcal{L}_{\mathrm{APSD}}^{\texttt{sg}}
&= \nabla_\theta D_{\mathrm{KL}}\! \left( \pi_\theta \middle\| \texttt{sg}\!\left[\pi_{\theta, \alpha}^{\mathrm{APSD}}\right] \right) \\ 
&= \alpha\nabla_\theta D_{\mathrm{KL}} \left(\pi_\theta\middle\|\pi\right) \\
&= \alpha\nabla_\theta\mathcal{L}_{\mathrm{SDPO}}(\theta).
\end{aligned}
\end{equation}
Consequently, full-target detachment degenerates the optimization into a uniformly rescaled SDPO update, completely discarding the R\'enyi-induced coupling term $-(1-\alpha)\nabla_\theta D_\alpha(\pi \parallel \pi_\theta)$. See Appendix~\ref{app:gradient} for the proof.
\end{proposition}

Proposition~\ref{proposition3} shows that detaching the entire target removes the geometric asymmetry established in Propositions~\ref {proposition1} and~\ref{proposition2}, reducing $\alpha$ to rescaling the teacher-directed gradient rather than reshaping the optimization trajectory.
To understand the role of the discarded gradient component, consider its contribution at the logit level, which is proportional to $(1-\alpha)
[\pi_{\theta,\alpha}^{\mathrm{APSD}}(v)-\pi_\theta(v)]$. For correct rollouts, where $\alpha_{\mathrm{corr}}\in(0,1)$, this contribution counteracts shifts toward the interpolated target. Under gradient descent, it raises the logits of tokens where $\pi_{\theta,\alpha}^{\mathrm{APSD}}(v)<\pi_\theta(v)$ and lowers those where $\pi_{\theta,\alpha}^{\mathrm{APSD}}(v)>\pi_\theta(v)$. This counteracting effect helps preserve successful student behaviors and prevents excessive changes induced by teacher.

In contrast, for incorrect rollouts, where $\alpha_{\mathrm{incorr}}>1$, the sign of this contribution reverses.
Gradient descent raises the logits of tokens for which $\pi_{\theta,\alpha}^{\mathrm{APSD}}(v)>\pi_\theta(v)$ and lowers those for which $\pi_{\theta,\alpha}^{\mathrm{APSD}}(v)<\pi_\theta(v)$, reinforcing shifts toward the extrapolated target. This strengthens
correction away from unsuccessful behaviors rather than merely rescaling the SDPO update.

These observations motivate APSD's selective stop-gradient design: detaching only the self-teacher preserves the target's dependence on the student and the resulting gradient contribution $-(1-\alpha)\nabla_\theta D_\alpha(\pi\parallel\pi_\theta)$. Maintaining this gradient flow is essential to APSD, allowing $\alpha$ to shape not only the magnitude but also the direction of policy updates according to rollout correctness.





\section{Experiment}
\begin{table*}[t]
\small
\centering
\renewcommand{\arraystretch}{0.9}

\definecolor{BaseGray}{RGB}{244,244,244}
\definecolor{LCBTopOne}{RGB}{244,190,205}
\definecolor{LCBTopTwo}{RGB}{252,229,235}

\makebox[\linewidth][c]{%
\begin{tabular}{
    l!{\vrule width 0.4pt}
    *{6}{>{\centering\arraybackslash}m{1.45cm}}
}
\toprule

\multirow{2}{*}{\textbf{Methods}}
& \multicolumn{2}{c}{\textbf{Qwen3-1.7B}}
& \multicolumn{2}{c}{\textbf{Qwen3-4B}}
& \multicolumn{2}{c}{\textbf{Qwen3-8B}} \\

\cmidrule(lr){2-3}
\cmidrule(lr){4-5}
\cmidrule(lr){6-7}

& \textbf{Avg@4}
& \textbf{Pass@4}
& \textbf{Avg@4}
& \textbf{Pass@4}
& \textbf{Avg@4}
& \textbf{Pass@4} \\

\midrule

\rowcolor{gray!20}
\textbf{Base}
& 15.84 & 20.61
& 25.00 & 32.82
& 25.57 & 35.11 \\

+ GRPO
& 30.34 & 33.59
& 37.98 & 41.22
& 40.27 & 45.80 \\

+ SDPO
& 32.06 & 32.06
& 42.56 & 42.75
& 45.04 & 47.33 \\

\midrule

+ SRPO
& \cellcolor{LCBTopTwo}\underline{33.78}
& \cellcolor{LCBTopOne}\textbf{35.11}
& 45.42
& 46.56
& 45.99
& 46.56 \\

+ RLSD
& 28.24
& 32.06
& 38.36
& 41.98
& 41.79
& 45.04 \\

\midrule

+ DistiLLM
& 30.34 & 31.30
& 44.85 & 45.80
& 45.99 & 48.09 \\

+ TAID
& 31.30 & 32.06
& 43.89 & 44.27
&  \cellcolor{LCBTopTwo}\underline{47.71} & \cellcolor{LCBTopTwo}\underline{48.85} \\

+ DemoPSD
& 31.49 & 32.06
& \cellcolor{LCBTopTwo}\underline{45.80} & \cellcolor{LCBTopTwo}\underline{47.33}
& 45.99 & 47.33 \\

\midrule

\textbf{+ APSD~(Ours)}
& \cellcolor{LCBTopOne}\textbf{34.16}
& \cellcolor{LCBTopTwo}\underline{34.35}
& \cellcolor{LCBTopOne}\textbf{47.71}
& \cellcolor{LCBTopOne}\textbf{48.85}
& \cellcolor{LCBTopOne}\textbf{49.62}
& \cellcolor{LCBTopOne}\textbf{51.15} \\

\bottomrule
\end{tabular}%
}
\vspace{-0.5em}
\caption{
RLRF performance on LCBv6 across Qwen3 model scales. Avg@4 denotes the average success
rate over four sampled solutions and Pass@4 denotes the fraction of
problems solved by at least one of them.
The best and second-best results in each column are highlighted in
{\setlength{\fboxsep}{1pt}\colorbox{LCBTopOne}{\textbf{dark pink}}}
and
{\setlength{\fboxsep}{1pt}\colorbox{LCBTopTwo}{\underline{light pink}}},
respectively.
}
\vspace{-0.8em}
\label{tab:rlrf_table}
\end{table*}

\begin{figure*}[t]
    \centering

    \begin{subfigure}[t]{0.49\textwidth}
        \centering
        \includegraphics[width=\linewidth]{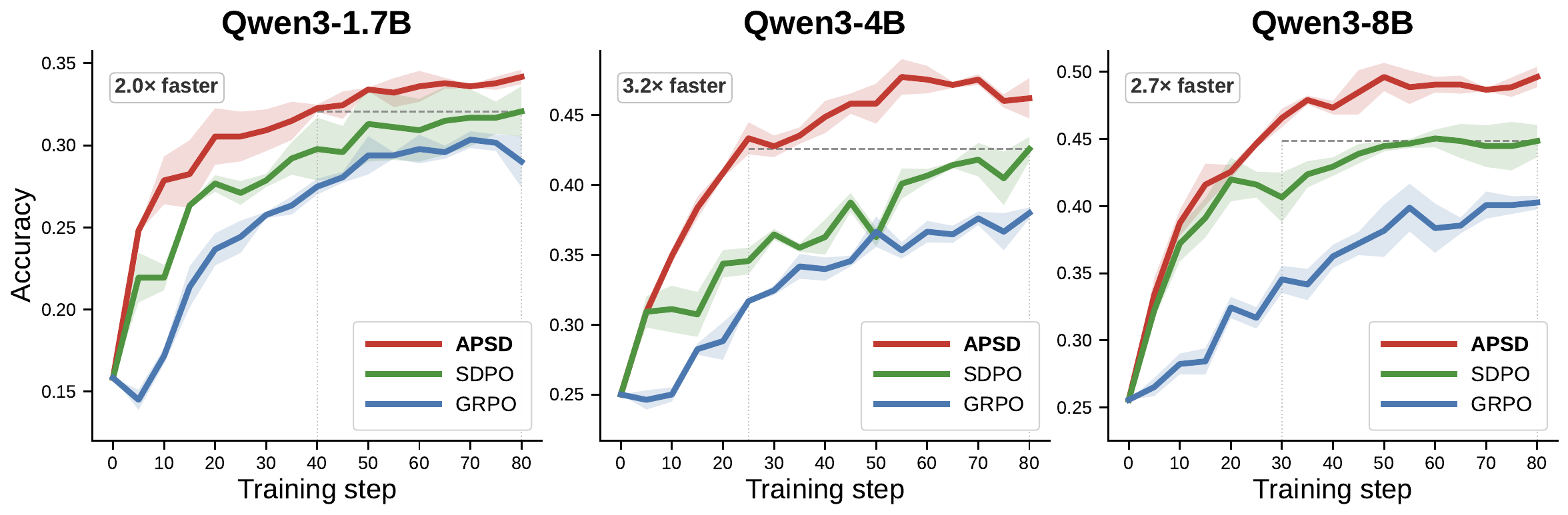}
        \vspace{-1.6em}

        \caption{Training dynamics}
        \label{fig:lcb-training-dynamics}
    \end{subfigure}
    \hfill
    \begin{subfigure}[t]{0.49\textwidth}
        \centering
        \includegraphics[width=\linewidth]{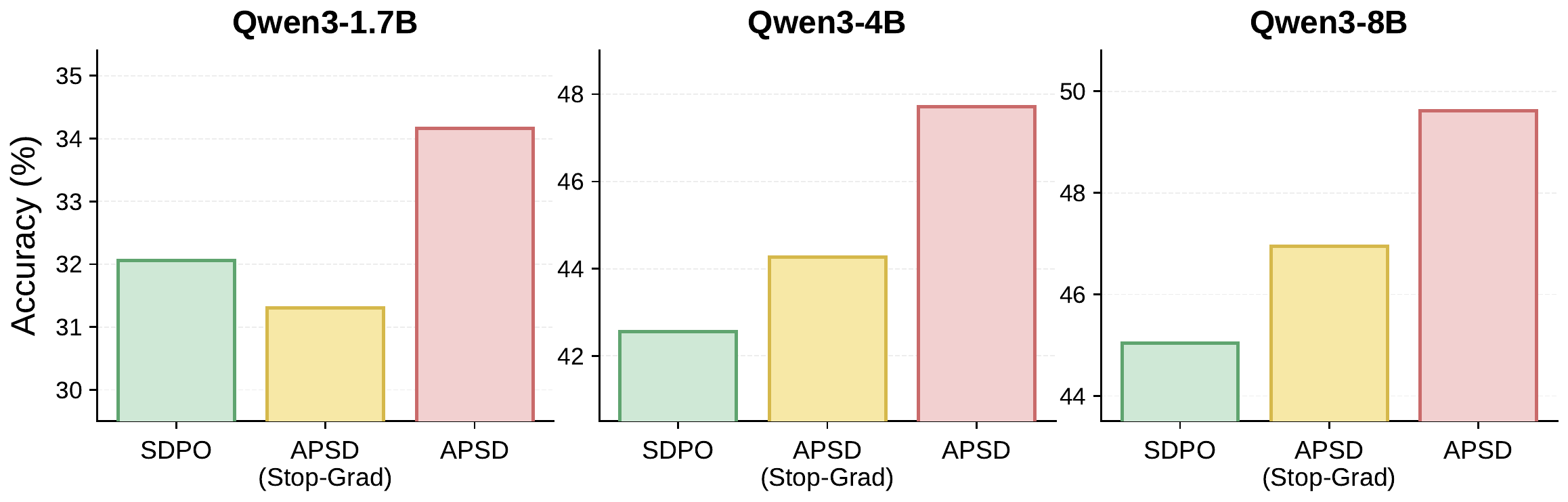}
        \vspace{-1.6em}

        \caption{Effect of target gradient flow}
        \label{fig:no-stop-gradient}
    \end{subfigure}
\vspace{-0.5em}

    \caption{
    Training dynamics and target gradient-flow ablation on LCBv6
    across Qwen3 model scales under RLRF.
    \textbf{(a)} Accuracy of APSD, SDPO, and GRPO over training steps.
    Shaded regions indicate the standard deviation across three random seeds.
    \textbf{(b)} Accuracy of SDPO, APSD with a detached target
    (APSD Stop-Grad), and full APSD.
    }
    \label{fig:training-and-gradient}
    \vspace{-1.4em}
\end{figure*}

\vspace{-0.2em}

\subsection{Experimental Setup}


\label{sec:experimental-setup}

We evaluate APSD under two complementary reinforcement learning settings
following prior self-distillation work~\citep{hubotter2026reinforcement}:
reinforcement learning with verifiable rewards (RLVR) and reinforcement learning with rich feedback (RLRF).
RLVR provides scalar outcome-level rewards, whereas RLRF additionally provides textual environment feedback that offers more detailed information about the generated solution.
\paragraph{RLVR.}
Following prior work~\citep{hubotter2026reinforcement}, we evaluate
APSD on five different benchmarks covering scientific reasoning and tool use.
For scientific reasoning, we use the Chemistry, Physics, Biology, and
Materials Science subsets of SciKnowEval~\citep{feng2025sciknowevalevaluatingmultilevelscientific};
for tool use, we use ToolAlpaca~\citep{tang2023toolalpacageneralizedtoollearning}.
In RLVR, where only scalar rewards are available, the self-teacher
is conditioned on a successful rollout from the same rollout group when
available, following SDPO.
Our experiments use Olmo3-7B-Instruct~\citep{olmo2026olmo3} and
Qwen3-8B~\citep{yang2025qwen3technicalreport} as base models.
Following the reporting protocol of prior work~\citep{hubotter2026reinforcement},
we report the highest Avg@16 accuracy achieved within the 5-hour wall-clock
training budget.





\paragraph{RLRF.}
Following the RLRF setup of SDPO \citep{hubotter2026reinforcement},
we evaluate APSD on LiveCodeBench~v6 (LCBv6)~\citep{jain2025livecodebench} with
Qwen3-1.7B, Qwen3-4B, and Qwen3-8B.
In RLRF, code execution feedback, including runtime errors and failed test cases,
additionally conditions the self-teacher during training.
We train all methods for 80 optimization steps and evaluate performance
using Avg@4 and Pass@4.

\paragraph{Baselines.}
We compare APSD with reinforcement learning and self-distillation baselines.
Our common baselines include GRPO~\citep{shao2024deepseekmathpushinglimitsmathematical},
SDPO~\citep{hubotter2026reinforcement}, SRPO~\citep{li2026unifying}, and RLSD~\citep{yang2026selfdistilledrlvr}.
For RLVR, we evaluate both off- and on-policy variants of GRPO.
For RLRF, we additionally consider DistiLLM~\citep{ko2024distillm},
TAID~\citep{shing2025taid}, and DemoPSD~\citep{li2026demopsddisagreementmodulatedpolicyselfdistillation}, adapting their
distribution-mixing strategies to the self-distillation setting.
Additional configurations and implementation
details are provided in Appendix~\ref{app:experimental-details}.

\subsection{Main Results}
\label{sec:main-results}
We first evaluate APSD on five RLVR benchmarks.
As shown in Table~\ref{tab:rlvr_table}, APSD achieves the highest average performance, while remaining competitive across individual tasks.
Importantly, APSD consistently outperforms SRPO and RLSD, which also optimize policy asymmetrically using  rollout correctness.
These results suggest that constructing asymmetric distributions based on
rollout correctness
 provides a more effective policy
optimization signal than other asymmetric formulations.

Under RLRF, we evaluate APSD on LCBv6 against correctness-guided self-distillation and distribution-mixing baselines across model scales. 
Consistent with the RLVR results, Table~\ref{tab:rlrf_table}
shows that APSD achieves the strongest overall performance, consistently outperforming SRPO and RLSD
and further supporting the effectiveness of correctness-dependent
interpolation and extrapolation under RLRF.
APSD also outperforms DistiLLM, TAID, and DemoPSD, whose distribution-mixing strategies are adapted to the SDPO.
These results highlight the benefit of incorporating rollout correctness into distillation target construction, beyond conventional distribution-mixing strategies.

\subsection{Training Efficiency}

We analyze the training dynamics of APSD across Qwen3 model scales.
Figure~\ref{fig:lcb-training-dynamics} shows that, compared with SDPO, APSD achieves a $2.6\times$ average reduction in the number of training steps required to reach the same performance level across model scales.
This improvement is consistent with APSD's target construction, which uses a student trust region to adjust the target's position relative to the student: targets remain closer to the student for correct rollouts and move farther away for incorrect ones.
APSD thus achieves both higher final performance and greater training-step efficiency.

\begin{table}[t]
\definecolor{LCBTopOne}{RGB}{244,190,205}
\centering
\small
\renewcommand{\arraystretch}{1}
\setlength{\tabcolsep}{5pt}
\begin{tabular}{lcc|cc}
\toprule
\multirow{2}{*}{\makecell[c]{\textbf{Sampling}\\\textbf{Distribution}}}
& \multicolumn{2}{c|}{\textbf{Correct rollout}}
& \multicolumn{2}{c}{\textbf{Incorrect rollout}} \\
& \textbf{Acc.}
& $\boldsymbol{\Delta}$
& \textbf{Acc.}
& $\boldsymbol{\Delta}$ \\
\midrule

Student 
& 78.57 & --
& 1.79 & -- \\

Inter. ($\pi^\text{APSD}_{\theta,0.5}$)
& \cellcolor{LCBTopOne}\textbf{83.93} & \cellcolor{LCBTopOne}\textbf{+5.36}
& 5.36 & +3.57 \\

Teacher
& 78.57 & +0.00
& 19.64 & +17.85 \\

Extra. ($\pi^\text{APSD}_{\theta,1.5}$)
& 75.00 & $-$3.57
& \cellcolor{LCBTopOne}\textbf{23.21} & \cellcolor{LCBTopOne}\textbf{+21.42} \\

\bottomrule
\end{tabular}

\caption{
Execution accuracy (\%) under different sampling distributions.
$\Delta$ denotes the percentage-point change relative to the student distribution.
}

\label{tab:correctness_alpha}
\end{table}

\begin{table}[t]
\centering
\small
\definecolor{LCBTopOne}{RGB}{244,190,205}
\setlength{\tabcolsep}{5pt}
\renewcommand{\arraystretch}{0.9}
\begin{tabular}{lccc}
\toprule
\textbf{Method}
& \textbf{Qwen3-1.7B}
& \textbf{Qwen3-4B}
& \textbf{Qwen3-8B} \\
\midrule
GRPO
& 30.34
& 37.98
& 40.27 \\
SDPO
& 32.06
& 42.56
& 45.04 \\
\midrule
APSD-C
& 33.59
& 46.18
& 48.85 \\
APSD-I
& 32.82
& 45.61
& 48.47 \\
\textbf{APSD}
& \cellcolor{LCBTopOne}\textbf{34.16}
& \cellcolor{LCBTopOne}\textbf{47.71}
& \cellcolor{LCBTopOne}\textbf{49.62} \\
\bottomrule
\end{tabular}
\caption{
Ablation of the correctness-specific components of APSD on LCBv6 (Avg@4).
}
\label{tab:asymmetry-ablation}
\end{table}

\subsection{Correctness-Dependent $\alpha$ Control}
\label{sub:sampling}
We conduct an additional experiment on LCBv6 to examine
the preferred APSD coefficient $\alpha$ by rollout correctness.
With a frozen Qwen3-8B student and self-teacher, we fix the first half
of each response and regenerate the suffix using APSD targets:
$\alpha=0$ (student), $0<\alpha<1$ (interpolation),
$\alpha=1$ (self-teacher), and $\alpha>1$ (extrapolation).

Table~\ref{tab:correctness_alpha} shows contrasts
between correct and incorrect rollouts.
For correct rollouts, interpolation improves execution accuracy
over the already effective student distribution, whereas the
self-teacher alone provides no gain and extrapolation reduces
performance. This suggests successful student behavior
should be preserved rather than aggressively overwritten.
For incorrect rollouts, the student rarely recovers a correct
solution, while the self-teacher substantially improves accuracy.
Extrapolation further improves performance beyond the self-teacher,
achieving the highest accuracy.
These results support APSD's correctness-dependent $\alpha$ selection.

\subsection{Ablation Study}
\label{sub:ablation}

\paragraph{Effect of Target Gradient Flow.}
To examine the role of gradient flow through the  target, we construct
a stop-gradient variant of APSD.
As shown in Figure~\ref{fig:no-stop-gradient}, 
the stop-gradient variant performs worse than the original APSD across
all Qwen3 model scales.
This result demonstrates the importance of preserving gradient flow
through the student-dependent target and is consistent with the analysis
in Section~\ref{sec:theoretical-analysis}, which identifies an additional
gradient contribution from the student--target coupling.

\paragraph{Effects of Interpolation and Extrapolation.}
We further assess the contribution of each correctness-specific operation
in APSD.
Table~\ref{tab:asymmetry-ablation} reports two partial variants:
APSD-C applies interpolation only to correct rollouts, while APSD-I
applies extrapolation only to incorrect rollouts.
Both variants improve over SDPO across all model scales, but neither
matches the performance of the full APSD.
These results indicate that the two operations are complementary, with
interpolation preserving successful behavior and extrapolation providing stronger correction for unsuccessful rollouts.

\begin{figure}[t]
    \centering
    \includegraphics[width=\columnwidth]{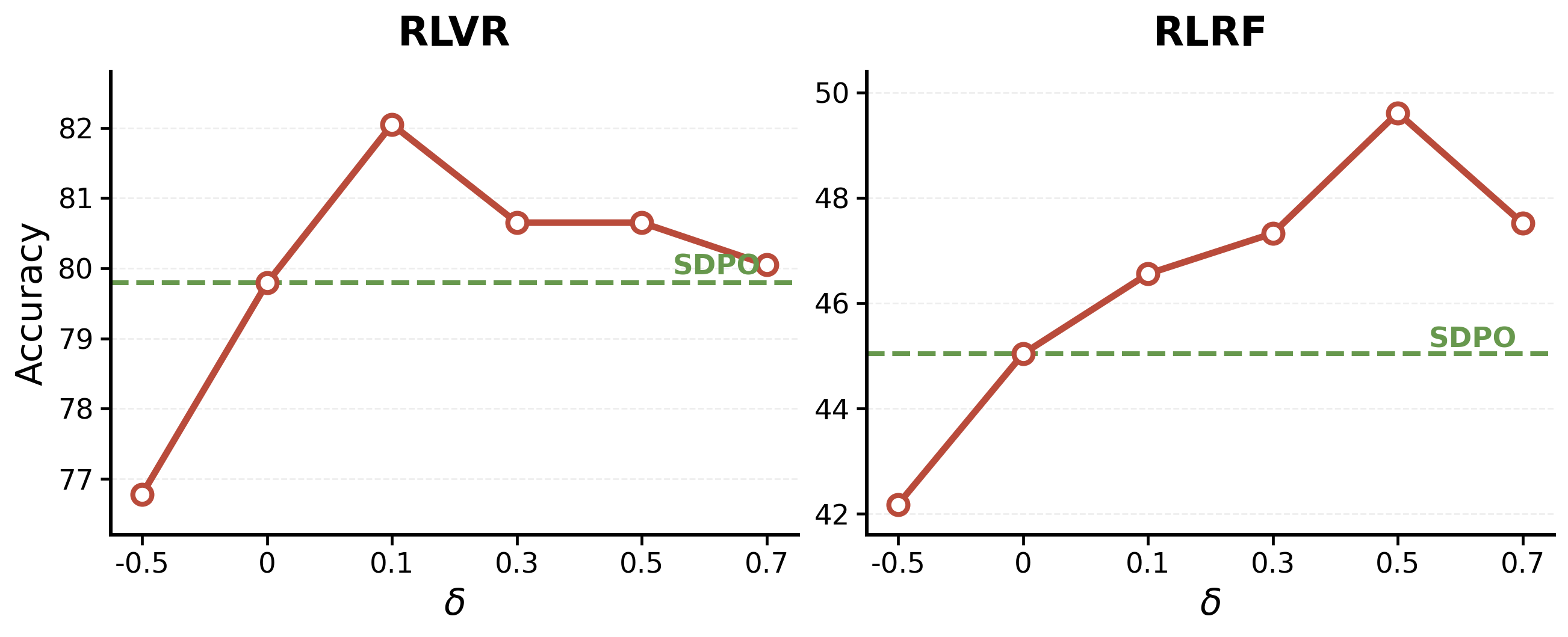}
    \vspace{-2.1em}
\caption{
Sensitivity of APSD to the asymmetry strength
$\delta$ under RLVR and RLRF. 
}
\vspace{-1.5em}
    \label{fig:alpha-sensitivity}
\end{figure}

\section{Sensitivity Study}
\label{app:alpha-sensitivity}


We examine APSD's sensitivity to asymmetry strength $\delta$ by setting
$\alpha_{\mathrm{corr}}=1-\delta$ and
$\alpha_{\mathrm{incorr}}=1+\delta$ around the SDPO configuration ($\delta=0$).
As shown in Figure~\ref{fig:alpha-sensitivity},
reversing the proposed asymmetry ($\delta<0$), such that correct
rollouts are extrapolated while incorrect rollouts are interpolated, degrades performance below SDPO in both RLVR and RLRF.
This result further confirms that the direction of the asymmetric adjustment
is critical.
For positive $\delta$, the preferred degree of asymmetry
differs across feedback settings: RLVR favors relatively
mild asymmetry, whereas RLRF favors stronger asymmetry.
Nevertheless, APSD consistently outperforms SDPO
across a broad range of positive $\delta$ values,
demonstrating robustness to the degree of asymmetry.

\section{Conclusion}

We present APSD, a self-distillation method that constructs asymmetric targets based on rollout correctness. APSD interpolates between student and self-teacher distributions to preserve successful behaviors and extrapolates beyond the self-teacher to strengthen correction of unsuccessful ones. Our analyses characterize APSD through divergence decomposition, reverse-KL barycentric geometry, and gradient flow through student-dependent targets. Experiments across RLVR and RLRF show strong overall performance against reinforcement-learning and self-distillation baselines
, supporting the need for correctness-dependent student–self-teacher distribution mixing.

\section*{Limitations}
Although APSD demonstrates consistent improvements across RLVR and RLRF settings, our study has a few limitations. Our experiments cover diverse tasks, including scientific reasoning, tool use, and code generation, while evaluation on additional language generation tasks is left for future work. Furthermore, due to computational resource constraints, we evaluate APSD on models with up to 8B parameters, and its effectiveness on substantially larger models remains to be investigated. Nevertheless, the consistent improvements across multiple model scales and feedback settings suggest the potential of correctness-dependent target construction for LLM post-training.

\bibliography{custom}

@book{boyd2004convex,
  title={Convex Optimization},
  author={Boyd, S.P. and Vandenberghe, L.},
  number={pt. 1},
  isbn={9780521833783},
  lccn={03063284},
  series={Berichte {\"u}ber verteilte messysteme},
  url={https://books.google.co.kr/books?id=xKAPkAEACAAJ},
  year={2004},
  publisher={Cambridge University Press}
}

@inproceedings{schulman2015trust,
  title={Trust region policy optimization},
  author={Schulman, John and Levine, Sergey and Abbeel, Pieter and Jordan, Michael and Moritz, Philipp},
  booktitle={International conference on machine learning},
  pages={1889--1897},
  year={2015},
  organization={Pmlr}
}

@article{van2014renyi,
  title={R{\'e}nyi divergence and Kullback-Leibler divergence},
  author={Van Erven, Tim and Harremos, Peter},
  journal={IEEE transactions on information theory},
  volume={60},
  number={7},
  pages={3797--3820},
  year={2014},
  publisher={IEEE}
}

@inproceedings{qiu2025multimodal,
  title={Multimodal variational autoencoder: A barycentric view},
  author={Qiu, Peijie and Zhu, Wenhui and Kumar, Sayantan and Chen, Xiwen and Yang, Jin and Sun, Xiaotong and Razi, Abolfazl and Wang, Yalin and Sotiras, Aristeidis},
  booktitle={Proceedings of the AAAI Conference on Artificial Intelligence},
  volume={39},
  number={19},
  pages={20060--20068},
  year={2025}
}

@inproceedings{
hubotter2026reinforcement,
title={Reinforcement Learning via Self-Distillation},
author={Jonas H{\"u}botter and Frederike L{\"u}beck and Lejs Deen Behric and Anton Baumann and Marco Bagatella and Daniel Marta and Ido Hakimi and Idan Shenfeld and Thomas Kleine Buening and Carlos Guestrin and Andreas Krause},
booktitle={Forty-third International Conference on Machine Learning},
year={2026},
url={https://openreview.net/forum?id=QkfkxyRizZ}
}

@inproceedings{
li2026unifying,
title={Unifying Group-Relative and Self-Distillation Policy Optimization via Sample Routing},
author={Gengsheng Li and Tianyu Yang and Junfeng Fang and Mingyang Song and Mao Zheng and Haiyun Guo and Dan Zhang and Jinqiao Wang and Tat-Seng Chua},
booktitle={Third Conference on Language Modeling},
year={2026},
url={https://openreview.net/forum?id=P2OuWwZspP}
}

@misc{feng2025sciknowevalevaluatingmultilevelscientific,
      title={SciKnowEval: Evaluating Multi-level Scientific Knowledge of Large Language Models}, 
      author={Kehua Feng and Xinyi Shen and Weijie Wang and Xiang Zhuang and Yuqi Tang and Qiang Zhang and Keyan Ding},
      year={2025},
      eprint={2406.09098},
      archivePrefix={arXiv},
      primaryClass={cs.CL},
      url={https://arxiv.org/abs/2406.09098}, 
}

@misc{tang2023toolalpacageneralizedtoollearning,
      title={ToolAlpaca: Generalized Tool Learning for Language Models with 3000 Simulated Cases}, 
      author={Qiaoyu Tang and Ziliang Deng and Hongyu Lin and Xianpei Han and Qiao Liang and Boxi Cao and Le Sun},
      year={2023},
      eprint={2306.05301},
      archivePrefix={arXiv},
      primaryClass={cs.CL},
      url={https://arxiv.org/abs/2306.05301}, 
}

@inproceedings{
jain2025livecodebench,
title={LiveCodeBench: Holistic and Contamination Free Evaluation of Large Language Models for Code},
author={Naman Jain and King Han and Alex Gu and Wen-Ding Li and Fanjia Yan and Tianjun Zhang and Sida Wang and Armando Solar-Lezama and Koushik Sen and Ion Stoica},
booktitle={The Thirteenth International Conference on Learning Representations},
year={2025},
url={https://openreview.net/forum?id=chfJJYC3iL}
}

@misc{olmo2026olmo3,
      title={Olmo 3}, 
      author={Team Olmo and : and Allyson Ettinger and Amanda Bertsch and Bailey Kuehl and David Graham and David Heineman and Dirk Groeneveld and Faeze Brahman and Finbarr Timbers and Hamish Ivison and Jacob Morrison and Jake Poznanski and Kyle Lo and Luca Soldaini and Matt Jordan and Mayee Chen and Michael Noukhovitch and Nathan Lambert and Pete Walsh and Pradeep Dasigi and Robert Berry and Saumya Malik and Saurabh Shah and Scott Geng and Shane Arora and Shashank Gupta and Taira Anderson and Teng Xiao and Tyler Murray and Tyler Romero and Victoria Graf and Akari Asai and Akshita Bhagia and Alexander Wettig and Alisa Liu and Aman Rangapur and Chloe Anastasiades and Costa Huang and Dustin Schwenk and Harsh Trivedi and Ian Magnusson and Jaron Lochner and Jiacheng Liu and Lester James V. Miranda and Maarten Sap and Malia Morgan and Michael Schmitz and Michal Guerquin and Michael Wilson and Regan Huff and Ronan Le Bras and Rui Xin and Rulin Shao and Sam Skjonsberg and Shannon Zejiang Shen and Shuyue Stella Li and Tucker Wilde and Valentina Pyatkin and Will Merrill and Yapei Chang and Yuling Gu and Zhiyuan Zeng and Ashish Sabharwal and Luke Zettlemoyer and Pang Wei Koh and Ali Farhadi and Noah A. Smith and Hannaneh Hajishirzi},
      year={2026},
      eprint={2512.13961},
      archivePrefix={arXiv},
      primaryClass={cs.CL},
      url={https://arxiv.org/abs/2512.13961}, 
}

@misc{yang2025qwen3technicalreport,
      title={Qwen3 Technical Report}, 
      author={An Yang and Anfeng Li and Baosong Yang and Beichen Zhang and Binyuan Hui and Bo Zheng and Bowen Yu and Chang Gao and Chengen Huang and Chenxu Lv and Chujie Zheng and Dayiheng Liu and Fan Zhou and Fei Huang and Feng Hu and Hao Ge and Haoran Wei and Huan Lin and Jialong Tang and Jian Yang and Jianhong Tu and Jianwei Zhang and Jianxin Yang and Jiaxi Yang and Jing Zhou and Jingren Zhou and Junyang Lin and Kai Dang and Keqin Bao and Kexin Yang and Le Yu and Lianghao Deng and Mei Li and Mingfeng Xue and Mingze Li and Pei Zhang and Peng Wang and Qin Zhu and Rui Men and Ruize Gao and Shixuan Liu and Shuang Luo and Tianhao Li and Tianyi Tang and Wenbiao Yin and Xingzhang Ren and Xinyu Wang and Xinyu Zhang and Xuancheng Ren and Yang Fan and Yang Su and Yichang Zhang and Yinger Zhang and Yu Wan and Yuqiong Liu and Zekun Wang and Zeyu Cui and Zhenru Zhang and Zhipeng Zhou and Zihan Qiu},
      year={2025},
      eprint={2505.09388},
      archivePrefix={arXiv},
      primaryClass={cs.CL},
      url={https://arxiv.org/abs/2505.09388}, 
}

@misc{yang2026selfdistilledrlvr,
      title={Self-Distilled RLVR}, 
      author={Chenxu Yang and Chuanyu Qin and Qingyi Si and Minghui Chen and Naibin Gu and Dingyu Yao and Zheng Lin and Weiping Wang and Jiaqi Wang and Nan Duan},
      year={2026},
      eprint={2604.03128},
      archivePrefix={arXiv},
      primaryClass={cs.LG},
      url={https://arxiv.org/abs/2604.03128}, 
}

@misc{shao2024deepseekmathpushinglimitsmathematical,
      title={DeepSeekMath: Pushing the Limits of Mathematical Reasoning in Open Language Models}, 
      author={Zhihong Shao and Peiyi Wang and Qihao Zhu and Runxin Xu and Junxiao Song and Xiao Bi and Haowei Zhang and Mingchuan Zhang and Y. K. Li and Y. Wu and Daya Guo},
      year={2024},
      eprint={2402.03300},
      archivePrefix={arXiv},
      primaryClass={cs.CL},
      url={https://arxiv.org/abs/2402.03300}, 
}

@inproceedings{
ko2024distillm,
title={Disti{LLM}: Towards Streamlined Distillation for Large Language Models},
author={Jongwoo Ko and Sungnyun Kim and Tianyi Chen and Se-Young Yun},
booktitle={Forty-first International Conference on Machine Learning},
year={2024},
url={https://openreview.net/forum?id=lsHZNNoC7r}
}

@inproceedings{shing2025taid,
  title={Taid: Temporally adaptive interpolated distillation for efficient knowledge transfer in language models},
  author={Shing, Makoto and Misaki, Kou and Bao, Han and Yokoi, Sho and Akiba, Takuya},
  booktitle={International Conference on Learning Representations},
  volume={2025},
  pages={92233--92256},
  year={2025}
}

@inproceedings{
shin2026amid,
title={{AM}iD: Knowledge Distillation for {LLM}s with \${\textbackslash}alpha\$-mixture Assistant Distribution},
author={Donghyeok Shin and Yeongmin Kim and Suhyeon Jo and Byeonghu Na and Il-chul Moon},
booktitle={The Fourteenth International Conference on Learning Representations},
year={2026},
url={https://openreview.net/forum?id=7WPJ0EgPdW}
}

@misc{yang2026learningteachergeneralizedonpolicy,
      title={Learning beyond Teacher: Generalized On-Policy Distillation with Reward Extrapolation}, 
      author={Wenkai Yang and Weijie Liu and Ruobing Xie and Kai Yang and Saiyong Yang and Yankai Lin},
      year={2026},
      eprint={2602.12125},
      archivePrefix={arXiv},
      primaryClass={cs.LG},
      url={https://arxiv.org/abs/2602.12125}, 
}

@misc{xu2026betaopsdderivingpolicyoptimization,
      title={$\beta$-OPSD: Deriving with Policy Optimization, Training with Self-Distillation}, 
      author={Jiawei Xu and Minghui Liu and Juzheng Zhang and Tom Goldstein and Furong Huang},
      year={2026},
      eprint={2607.28582},
      archivePrefix={arXiv},
      primaryClass={cs.LG},
      url={https://arxiv.org/abs/2607.28582}, 
}

@misc{li2026demopsddisagreementmodulatedpolicyselfdistillation,
      title={DemoPSD: Disagreement-Modulated Policy Self-Distillation}, 
      author={Yunhe Li and Hao Shi and Wenhao Liu and Mengzhe Ruan and Hanxu Hou and Zhongxiang Dai and Shuang Qiu and Linqi Song},
      year={2026},
      eprint={2607.02502},
      archivePrefix={arXiv},
      primaryClass={cs.LG},
      url={https://arxiv.org/abs/2607.02502}, 
}

@misc{ye2026onpolicycontextdistillationlanguage,
      title={On-Policy Context Distillation for Language Models}, 
      author={Tianzhu Ye and Li Dong and Xun Wu and Shaohan Huang and Furu Wei},
      year={2026},
      eprint={2602.12275},
      archivePrefix={arXiv},
      primaryClass={cs.CL},
      url={https://arxiv.org/abs/2602.12275}, 
}

@inproceedings{
shenfeld2026selfdistillation,
title={Self-Distillation Enables Continual Learning},
author={Idan Shenfeld and Mehul Damani and Jonas H{\"u}botter and Pulkit Agrawal},
booktitle={Forty-third International Conference on Machine Learning},
year={2026},
url={https://openreview.net/forum?id=qA6FgH0nnZ}
}

@inproceedings{
zhao2026selfdistilled,
title={Self-Distilled Reasoner: On-Policy Self-Distillation for Large Language Models},
author={Siyan Zhao and Zhihui Xie and Mengchen Liu and Jing Huang and Guan Pang and Feiyu Chen and Aditya Grover},
booktitle={Forty-third International Conference on Machine Learning},
year={2026},
url={https://openreview.net/forum?id=Jpxfof0EaS}
}

@inproceedings{
kim2026rebellious,
title={Rebellious Student: Reversing Teacher Signals for Reasoning Exploration with Self-Distilled {RLVR}},
author={Jeonghye Kim and Jiwon Jeon and Dongsheng Li and Yuqing Yang},
booktitle={COLM 2026 Workshop on Efficient Reasoning},
year={2026},
url={https://openreview.net/forum?id=55DSgS3b8h}
}

@inproceedings{
song2026expanding,
title={Expanding the Capabilities of Reinforcement Learning via Text Feedback},
author={Yuda Song and Lili Chen and Fahim Tajwar and R{\'e}mi Munos and Deepak Pathak and Drew Bagnell and Aarti Singh and Andrea Zanette},
booktitle={Forty-third International Conference on Machine Learning},
year={2026},
url={https://openreview.net/forum?id=3voRLZLkwU}
}

@inproceedings{jung-etal-2025-todi,
    title = "{T}o{D}i: Token-wise Distillation via Fine-Grained Divergence Control",
    author = "Jung, Seongryong  and
      Yoon, Suwan  and
      Kim, DongGeon  and
      Lee, Hwanhee",
    editor = "Christodoulopoulos, Christos  and
      Chakraborty, Tanmoy  and
      Rose, Carolyn  and
      Peng, Violet",
    booktitle = "Proceedings of the 2025 Conference on Empirical Methods in Natural Language Processing",
    month = nov,
    year = "2025",
    address = "Suzhou, China",
    publisher = "Association for Computational Linguistics",
    url = "https://aclanthology.org/2025.emnlp-main.409/",
    doi = "10.18653/v1/2025.emnlp-main.409",
    pages = "8078--8091",
    ISBN = "979-8-89176-332-6"
}

@misc{hinton2015distillingknowledgeneuralnetwork,
      title={Distilling the Knowledge in a Neural Network}, 
      author={Geoffrey Hinton and Oriol Vinyals and Jeff Dean},
      year={2015},
      eprint={1503.02531},
      archivePrefix={arXiv},
      primaryClass={stat.ML},
      url={https://arxiv.org/abs/1503.02531}, 
}

@inproceedings{
gu2024minillm,
title={Mini{LLM}: Knowledge Distillation of Large Language Models},
author={Yuxian Gu and Li Dong and Furu Wei and Minlie Huang},
booktitle={The Twelfth International Conference on Learning Representations},
year={2024},
url={https://openreview.net/forum?id=5h0qf7IBZZ}
}

@article{Guo_2025,
   title={DeepSeek-R1 incentivizes reasoning in LLMs through reinforcement learning},
   volume={645},
   ISSN={1476-4687},
   url={http://dx.doi.org/10.1038/s41586-025-09422-z},
   DOI={10.1038/s41586-025-09422-z},
   number={8081},
   journal={Nature},
   publisher={Springer Science and Business Media LLC},
   author={Guo, Daya and Yang, Dejian and Zhang, Haowei and Song, Junxiao and Wang, Peiyi and Zhu, Qihao and Xu, Runxin and Zhang, Ruoyu and Ma, Shirong and Bi, Xiao and Zhang, Xiaokang and Yu, Xingkai and Wu, Yu and Wu, Z. F. and Gou, Zhibin and Shao, Zhihong and Li, Zhuoshu and Gao, Ziyi and Liu, Aixin and Xue, Bing and Wang, Bingxuan and Wu, Bochao and Feng, Bei and Lu, Chengda and Zhao, Chenggang and Deng, Chengqi and Ruan, Chong and Dai, Damai and Chen, Deli and Ji, Dongjie and Li, Erhang and Lin, Fangyun and Dai, Fucong and Luo, Fuli and Hao, Guangbo and Chen, Guanting and Li, Guowei and Zhang, H. and Xu, Hanwei and Ding, Honghui and Gao, Huazuo and Qu, Hui and Li, Hui and Guo, Jianzhong and Li, Jiashi and Chen, Jingchang and Yuan, Jingyang and Tu, Jinhao and Qiu, Junjie and Li, Junlong and Cai, J. L. and Ni, Jiaqi and Liang, Jian and Chen, Jin and Dong, Kai and Hu, Kai and You, Kaichao and Gao, Kaige and Guan, Kang and Huang, Kexin and Yu, Kuai and Wang, Lean and Zhang, Lecong and Zhao, Liang and Wang, Litong and Zhang, Liyue and Xu, Lei and Xia, Leyi and Zhang, Mingchuan and Zhang, Minghua and Tang, Minghui and Zhou, Mingxu and Li, Meng and Wang, Miaojun and Li, Mingming and Tian, Ning and Huang, Panpan and Zhang, Peng and Wang, Qiancheng and Chen, Qinyu and Du, Qiushi and Ge, Ruiqi and Zhang, Ruisong and Pan, Ruizhe and Wang, Runji and Chen, R. J. and Jin, R. L. and Chen, Ruyi and Lu, Shanghao and Zhou, Shangyan and Chen, Shanhuang and Ye, Shengfeng and Wang, Shiyu and Yu, Shuiping and Zhou, Shunfeng and Pan, Shuting and Li, S. S. and Zhou, Shuang and Wu, Shaoqing and Yun, Tao and Pei, Tian and Sun, Tianyu and Wang, T. and Zeng, Wangding and Liu, Wen and Liang, Wenfeng and Gao, Wenjun and Yu, Wenqin and Zhang, Wentao and Xiao, W. L. and An, Wei and Liu, Xiaodong and Wang, Xiaohan and Chen, Xiaokang and Nie, Xiaotao and Cheng, Xin and Liu, Xin and Xie, Xin and Liu, Xingchao and Yang, Xinyu and Li, Xinyuan and Su, Xuecheng and Lin, Xuheng and Li, X. Q. and Jin, Xiangyue and Shen, Xiaojin and Chen, Xiaosha and Sun, Xiaowen and Wang, Xiaoxiang and Song, Xinnan and Zhou, Xinyi and Wang, Xianzu and Shan, Xinxia and Li, Y. K. and Wang, Y. Q. and Wei, Y. X. and Zhang, Yang and Xu, Yanhong and Li, Yao and Zhao, Yao and Sun, Yaofeng and Wang, Yaohui and Yu, Yi and Zhang, Yichao and Shi, Yifan and Xiong, Yiliang and He, Ying and Piao, Yishi and Wang, Yisong and Tan, Yixuan and Ma, Yiyang and Liu, Yiyuan and Guo, Yongqiang and Ou, Yuan and Wang, Yuduan and Gong, Yue and Zou, Yuheng and He, Yujia and Xiong, Yunfan and Luo, Yuxiang and You, Yuxiang and Liu, Yuxuan and Zhou, Yuyang and Zhu, Y. X. and Huang, Yanping and Li, Yaohui and Zheng, Yi and Zhu, Yuchen and Ma, Yunxian and Tang, Ying and Zha, Yukun and Yan, Yuting and Ren, Z. Z. and Ren, Zehui and Sha, Zhangli and Fu, Zhe and Xu, Zhean and Xie, Zhenda and Zhang, Zhengyan and Hao, Zhewen and Ma, Zhicheng and Yan, Zhigang and Wu, Zhiyu and Gu, Zihui and Zhu, Zijia and Liu, Zijun and Li, Zilin and Xie, Ziwei and Song, Ziyang and Pan, Zizheng and Huang, Zhen and Xu, Zhipeng and Zhang, Zhongyu and Zhang, Zhen},
   year={2025},
   month=Sept, pages={633–638} }

@misc{kimiteam2025kimik15scalingreinforcement,
      title={Kimi k1.5: Scaling Reinforcement Learning with LLMs}, 
      author={Kimi Team and Angang Du and Bofei Gao and Bowei Xing and Changjiu Jiang and Cheng Chen and Cheng Li and Chenjun Xiao and Chenzhuang Du and Chonghua Liao and Chuning Tang and Congcong Wang and Dehao Zhang and Enming Yuan and Enzhe Lu and Fengxiang Tang and Flood Sung and Guangda Wei and Guokun Lai and Haiqing Guo and Han Zhu and Hao Ding and Hao Hu and Hao Yang and Hao Zhang and Haotian Yao and Haotian Zhao and Haoyu Lu and Haoze Li and Haozhen Yu and Hongcheng Gao and Huabin Zheng and Huan Yuan and Jia Chen and Jianhang Guo and Jianlin Su and Jianzhou Wang and Jie Zhao and Jin Zhang and Jingyuan Liu and Junjie Yan and Junyan Wu and Lidong Shi and Ling Ye and Longhui Yu and Mengnan Dong and Neo Zhang and Ningchen Ma and Qiwei Pan and Qucheng Gong and Shaowei Liu and Shengling Ma and Shupeng Wei and Sihan Cao and Siying Huang and Tao Jiang and Weihao Gao and Weimin Xiong and Weiran He and Weixiao Huang and Weixin Xu and Wenhao Wu and Wenyang He and Xianghui Wei and Xianqing Jia and Xingzhe Wu and Xinran Xu and Xinxing Zu and Xinyu Zhou and Xuehai Pan and Y. Charles and Yang Li and Yangyang Hu and Yangyang Liu and Yanru Chen and Yejie Wang and Yibo Liu and Yidao Qin and Yifeng Liu and Ying Yang and Yiping Bao and Yulun Du and Yuxin Wu and Yuzhi Wang and Zaida Zhou and Zhaoji Wang and Zhaowei Li and Zhen Zhu and Zheng Zhang and Zhexu Wang and Zhilin Yang and Zhiqi Huang and Zihao Huang and Ziyao Xu and Zonghan Yang and Zongyu Lin},
      year={2025},
      eprint={2501.12599},
      archivePrefix={arXiv},
      primaryClass={cs.AI},
      url={https://arxiv.org/abs/2501.12599}, 
}

@misc{xing2026trustregiononpolicydistillation,
      title={Trust Region On-Policy Distillation}, 
      author={Xingrun Xing and Haoqing Wang and Boyan Gao and Ziheng Li and Yehui Tang},
      year={2026},
      eprint={2606.01249},
      archivePrefix={arXiv},
      primaryClass={cs.LG},
      url={https://arxiv.org/abs/2606.01249}, 
}
\clearpage
\newpage

\appendix

\label{sec:appendix}



\paragraph{The Use Of Large Language Model Assistants}
We drafted the manuscript ourselves, and  GPT-5.6 Sol and GPT-6 Astra were used only for overall language refinement.

\begin{figure*}[!t]
    \centering

    \begin{subfigure}[t]{0.48\textwidth}
        \centering
        \includegraphics[width=\linewidth]{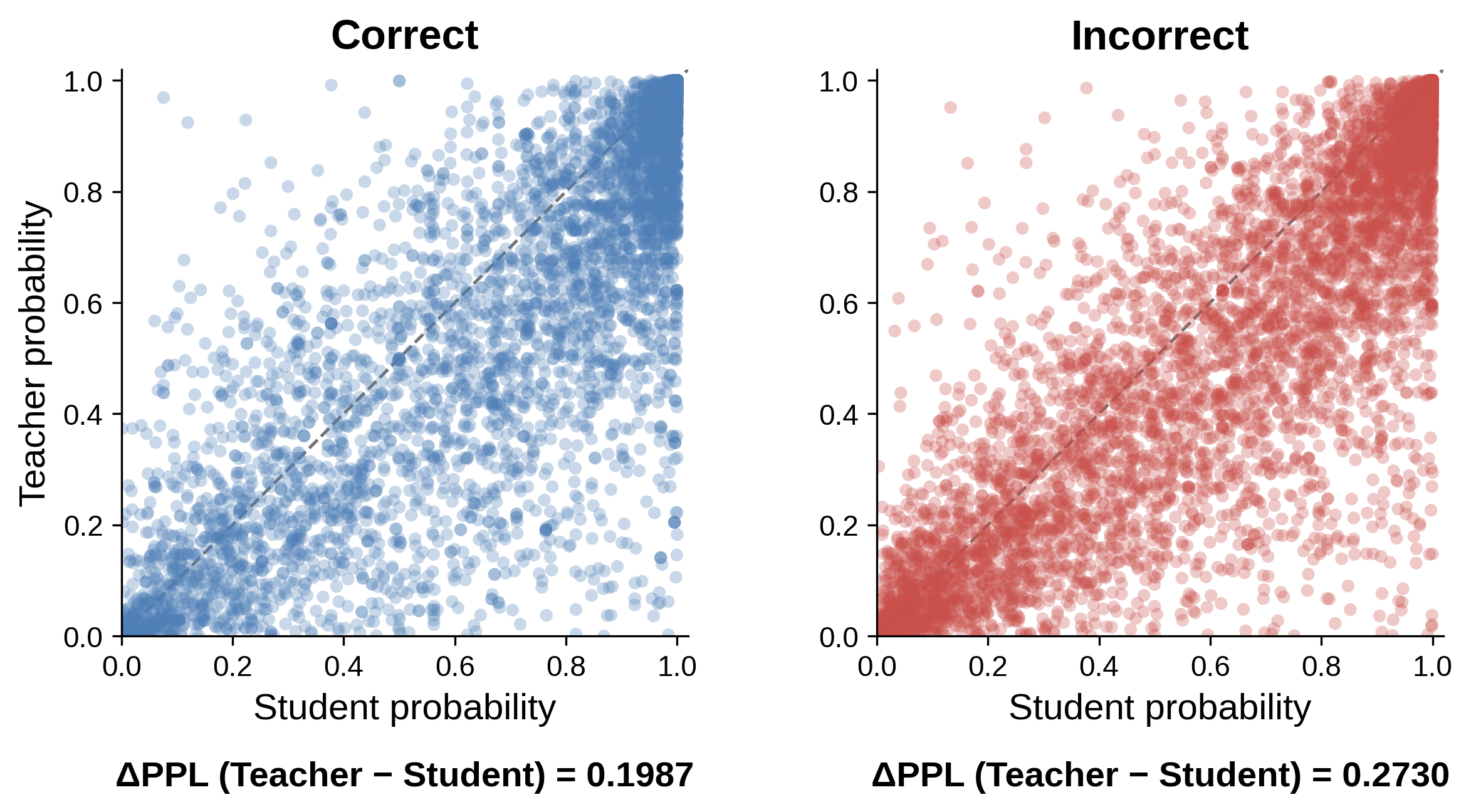}
        \caption{External OPD teacher.}
        \label{fig:opd_scatter}
    \end{subfigure}
    \hfill
    \begin{subfigure}[t]{0.48\textwidth}
        \centering
        \includegraphics[width=\linewidth]{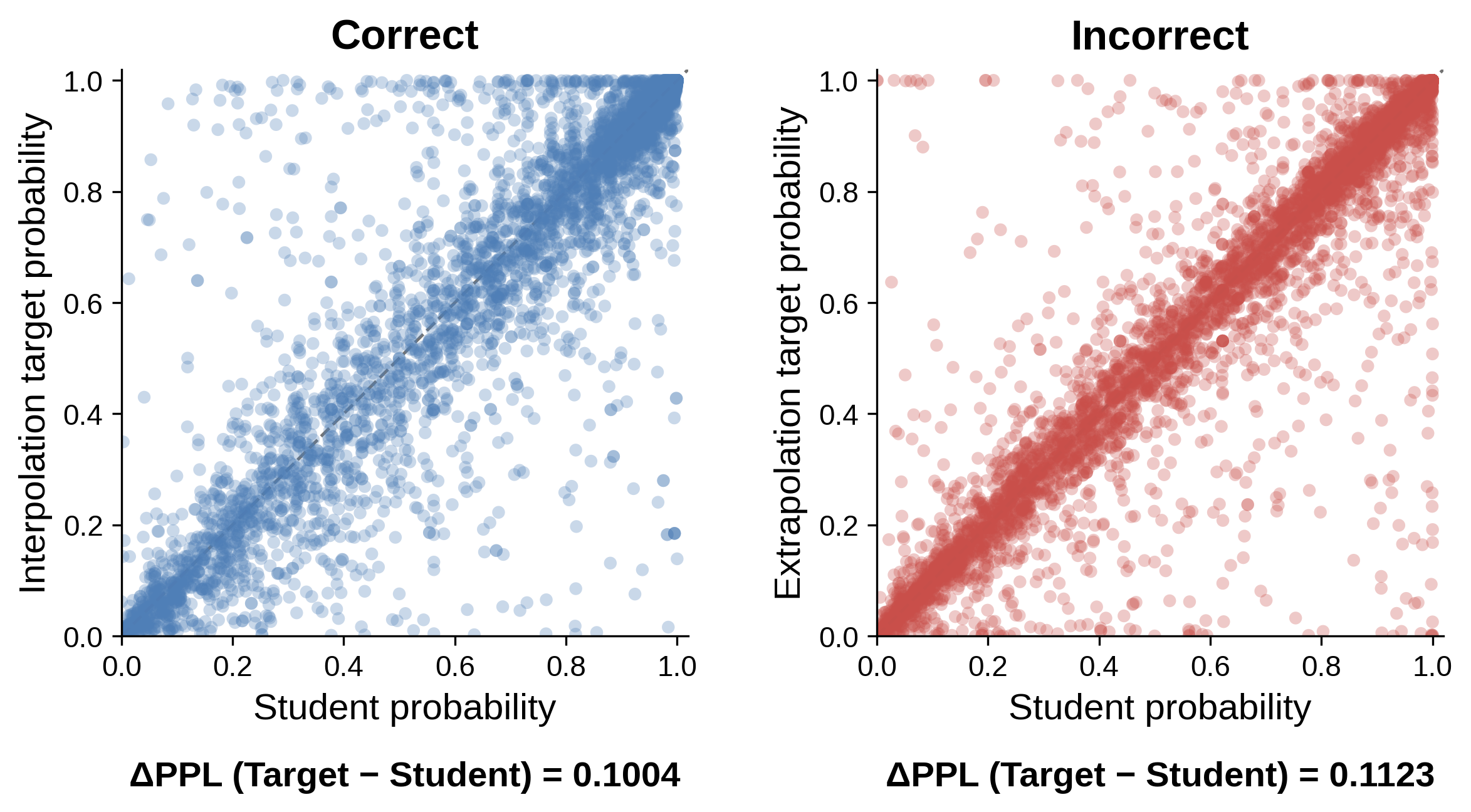}
        \caption{Correctness-dependent APSD target.}
        \label{fig:apsd_target_scatter}
    \end{subfigure}

    \caption{
    Additional analysis of teacher--student discrepancy on LiveCodeBench v6.
    \textbf{(a)} Qwen3-8B student versus a separate Qwen3-32B OPD teacher.
    The teacher--student perplexity gap is larger for incorrect rollouts than
    for correct rollouts.
    \textbf{(b)} Correctness-dependent APSD targets constructed from the
    SDPO self-teacher.
    Interpolation reduces the discrepancy for correct rollouts, whereas
    extrapolation increases it for incorrect rollouts.
    Each panel contains 10,000 sampled tokens.
    $\Delta$PPL denotes the mean perplexity difference over the corresponding
    rollouts.
    }
    \label{fig:additional_teacher_student_analysis}
\end{figure*}

\section{Additional Analysis of Teacher--Student Discrepancy}
\label{app:teacher_student_analysis}




\paragraph{Comparison with an External OPD Teacher.}
To examine whether the behavior observed in
Figure~\ref{fig:motivation} is specific to self-distillation,
we additionally compare the Qwen3-8B student with a larger Qwen3-32B
OPD teacher.
Unlike the SDPO self-teacher, the OPD teacher is a separate model and
receives the same original prompt as the student.

As shown in Figure~\ref{fig:additional_teacher_student_analysis}(a),
the external teacher exhibits a different correctness-dependent pattern.
The teacher--student perplexity gap is larger for incorrect rollouts than
for correct rollouts.
In contrast, for the SDPO self-teacher in
Figure~\ref{fig:motivation}, the gap is smaller on
incorrect rollouts than on correct rollouts.
This comparison highlights a characteristic difficulty of
self-distillation, as even after conditioning on feedback, the self-teacher
can remain close to the student distribution on failed trajectories,
limiting the strength of corrective supervision.

\paragraph{Effect of Correctness-Dependent Asymmetry.}
We next visualize how the proposed correctness-dependent transformation
changes the relationship between the student and self-teacher
distributions.
For correct rollouts, we interpolate between the student and self-teacher,
while for incorrect rollouts, we extrapolate toward the self-teacher
relative to the student.

As shown in Figure~\ref{fig:additional_teacher_student_analysis}(b),
these two operations adjust the discrepancy in opposite directions.
For correct rollouts, interpolation reduces the perplexity gap from
$0.2092$ for the original self-teacher to $0.1004$, keeping the target
closer to the already successful student behavior.
For incorrect rollouts, extrapolation increases the gap from $0.0720$
to $0.1123$, moving the target further away from the failed student
trajectory.
This behavior matches the intended role of APSD by applying conservative
distillation after success and stronger corrective separation after
failure.

\paragraph{{Qualitative Case Study of Asymmetric Targets}}
\label{app:token-level-analysis}

\begin{figure*}[!t]
    \centering
    \includegraphics[width=1.0\textwidth]
    {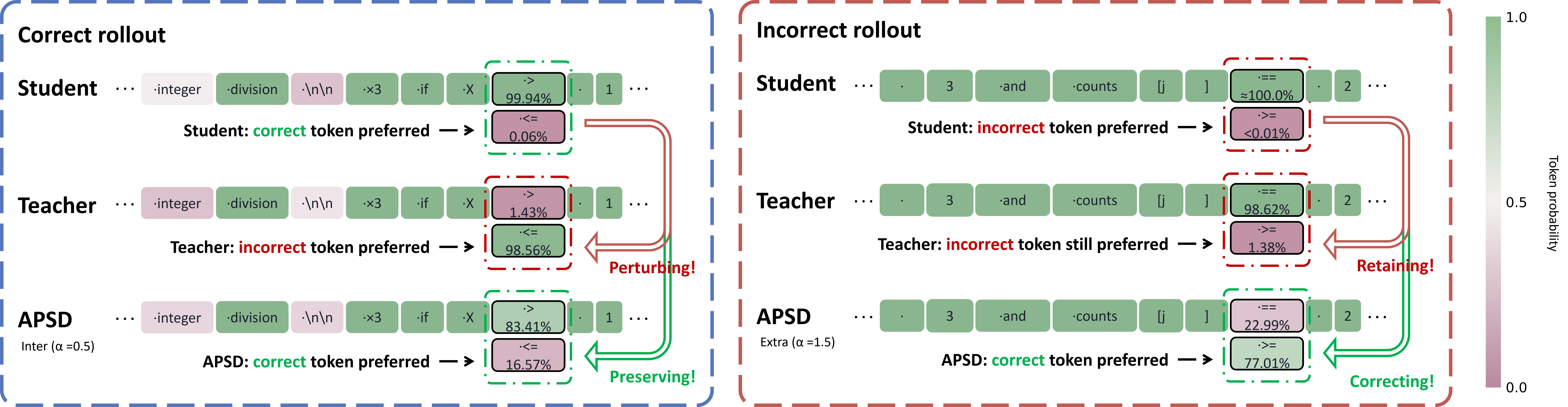}
    \caption{
    Token-level effect of APSD on representative correct and incorrect rollouts.
    For the correct rollout, interpolation shifts the target toward the student's
    successful choice, while for the incorrect rollout, extrapolation increases
    the probability of the corrective alternative.
    }
    \label{fig:apsd-token-examples}
\end{figure*}

Figure~\ref{fig:apsd-token-examples} illustrates how APSD modifies the
self-teacher distribution in the representative cases shown in
Figure~\ref{fig:motivation}.
For the correct rollout, interpolation moves the target back toward the
student's successful token choice.
For the incorrect rollout, extrapolation amplifies the teacher's corrective
direction and shifts probability toward the alternative associated with the
correct behavior.


\section{Proofs}
\label{app:proofs}

For readability, we omit all conditioning variables and token indices throughout the appendix, and denote the student, self-teacher, and PoE target distributions by $\pi_\theta$, $\pi$, and $\pi_{\theta, \alpha}^{\mathrm{APSD}}$, respectively. 

\subsection{Proof of Proposition 1}
\label{app:renyi_derivation}

\paragraph{Asymmetric divergence decomposition.}
Recall that $\pi_{\theta, \alpha}^{\mathrm{APSD}}$ is defined as
\begin{equation}
\label{eq:app_poe}
\begin{aligned}
\pi_{\theta, \alpha}^{\mathrm{APSD}}(v)
&=
\frac{
\pi_\theta(v)^{1-\alpha}
\pi(v)^\alpha
}{
Z_{\theta,\alpha}
}.
\end{aligned}
\end{equation}

The normalizing term is
\begin{equation}
\label{eq:app_normalizer}
\begin{aligned}
Z_{\theta,\alpha}
&=
\sum_{v\in V}
\pi_\theta(v)^{1-\alpha}
\pi(v)^\alpha.
\end{aligned}
\end{equation}

The APSD objective for a single decoding step is
\begin{equation}
\begin{aligned}
\mathcal{L}_{\mathrm{APSD}}
&=
D_{\mathrm{KL}}
\left(
\pi_\theta
\middle\|
\pi_{\theta, \alpha}^{\mathrm{APSD}}
\right)
\\
&=
\sum_{v\in V}
\pi_\theta(v)
\log
\frac{
\pi_\theta(v)
}{
\pi_{\theta, \alpha}^{\mathrm{APSD}}(v)
}.
\end{aligned}
\end{equation}

Substituting Eq.~\eqref{eq:app_poe} gives
\begin{equation}
\begin{aligned}
\mathcal{L}_{\mathrm{APSD}}
&=
\sum_{v\in V}
\pi_\theta(v)
\left[
\alpha
\log
\frac{
\pi_\theta(v)
}{
\pi(v)
}
+
\log Z_{\theta,\alpha}
\right]
\\
&=
\alpha
D_{\mathrm{KL}}
\left(
\pi_\theta
\middle\|
\pi
\right)
+
\log Z_{\theta,\alpha}.
\end{aligned}
\end{equation}

For $\alpha\neq1$, the R\'enyi divergence of order $\alpha$~\citep{van2014renyi} is defined as
\begin{equation}
\begin{aligned}
D_\alpha(P\|Q)
&=
\frac{1}{\alpha-1}
\log
\sum_{v\in V}
P(v)^\alpha
Q(v)^{1-\alpha}.
\end{aligned}
\end{equation}

Setting $P=\pi$ and $Q=\pi_\theta$ yields
\begin{equation}
\begin{aligned}
\log Z_{\theta,\alpha}
&=
(\alpha-1)
D_\alpha
\left(
\pi
\middle\|
\pi_\theta
\right)
\\
&=
-(1-\alpha)
D_\alpha
\left(
\pi
\middle\|
\pi_\theta
\right).
\end{aligned}
\end{equation}

Therefore,
\begin{equation}
\label{eq:app_renyi_decomposition}
\begin{aligned}
\mathcal{L}_{\mathrm{APSD}}
&=
\alpha
D_{\mathrm{KL}}
\left(
\pi_\theta
\middle\|
\pi
\right)
\\
&\quad
-
(1-\alpha)
D_\alpha
\left(
\pi
\middle\|
\pi_\theta
\right).
\end{aligned}
\end{equation}

When $\alpha=1$, $\pi_{\theta, \alpha}^{\mathrm{APSD}}$ reduces to the self-teacher distribution $\pi$, yielding $\mathcal{L}_{\mathrm{APSD}}=D_{\mathrm{KL}}(\pi_\theta\|\pi)$.\hfill $\square$

\subsection{Proof of Proposition 2}
\label{app:barycenter}

\paragraph{Characterization of $\pi_{\theta,\alpha}^{\mathrm{APSD}}$ as a reverse-KL barycenter.}
Following the derivation in \citet{qiu2025multimodal},
we rewrite the generalized reverse-KL barycentric objective using $(1-\alpha)+\alpha=1$, we have as

\begin{equation}
\begin{aligned}
&\pi_{\theta, \alpha}^{\mathrm{APSD}}
= \operatorname*{arg\,min}_{r\in\Delta(\mathcal{V})}
\Bigl[(1-\alpha)D_{\mathrm{KL}}(r\|\pi_\theta) \\
&\qquad\qquad\qquad\qquad\qquad\qquad
+\alpha D_{\mathrm{KL}}(r\|\pi)\Bigr]
\\[0.5ex]
&= \operatorname*{arg\,min}_{r\in\Delta(\mathcal{V})}
\sum_{v\in\mathcal{V}} r(v)
\Biggl[(1-\alpha)\log\frac{r(v)}{\pi_\theta(v)} \\
&\qquad\qquad\qquad\qquad\qquad\qquad
+\alpha\log\frac{r(v)}{\pi(v)}\Biggr]
\\[0.5ex]
&= \operatorname*{arg\,min}_{r\in\Delta(\mathcal{V})}
\sum_{v\in\mathcal{V}} r(v)
\log
\frac{r(v)}
{\pi_\theta(v)^{1-\alpha}\pi(v)^\alpha}
\\[0.5ex]
&= \operatorname*{arg\,min}_{r\in\Delta(\mathcal{V})}
\sum_{v\in\mathcal{V}} r(v)
\log
\frac{r(v)}
{Z_{\theta,\alpha}{\frac{
\pi_\theta(v)^{1-\alpha}\pi(v)^\alpha
}{
Z_{\theta,\alpha}
}}}
\\[0.5ex]
&= \operatorname*{arg\,min}_{r\in\Delta(\mathcal{V})}
\Bigl[
D_{\mathrm{KL}}\bigl(r\|{\frac{
\pi_\theta(v)^{1-\alpha}\pi(v)^\alpha
}{
Z_{\theta,\alpha}
}}\bigr)
-\log Z_{\theta,\alpha}
\Bigr]
\\[0.5ex]
&= \operatorname*{arg\,min}_{r\in\Delta(\mathcal{V})}
D_{\mathrm{KL}}\bigl(r\|{\frac{
\pi_\theta(v)^{1-\alpha}\pi(v)^\alpha
}{
Z_{\theta,\alpha}
}}\bigr)
\end{aligned}
\end{equation}

Since the KL divergence is minimized at equality, the unique minimizer is
\begin{equation}
\begin{aligned}
r(v)
=
\frac{
\pi_\theta(v)^{1-\alpha}\pi(v)^\alpha
}{
Z_{\theta,\alpha}
}
&=
\pi_{\theta, \alpha}^{\mathrm{APSD}}(v).
\end{aligned}
\end{equation}

Thus, $\alpha$ directly determines the relative weighting of the student and self-teacher in the generalized reverse-KL barycentric target. For $0<\alpha<1$, this corresponds to interpolation between the two distributions, while $\alpha>1$ extrapolates along the same normalized product form.
Unlike this reverse-KL barycenter, which takes a normalized
product-of-experts (PoE) form, the forward-KL barycenter takes
the mixture-of-experts (MoE) form
$(1-\alpha)\pi_\theta+\alpha\pi$.
For $\alpha>1$, the negative student coefficient can yield
negative probabilities, so this mixture does not generally
define a valid distribution.
In contrast, under the strict positivity assumption, the normalized
PoE form remains a valid distribution and supports extrapolation
beyond the self-teacher.

\paragraph{Monotonic growth of the student-centered trust-region
radius with $\alpha$.}
Using the derivation in \citep{hubotter2026reinforcement, xing2026trustregiononpolicydistillation}, we fix the student and self-teacher distributions $\pi_\theta$ and $\pi$,
and assume that both have full support over the finite vocabulary
$\mathcal{V}$ and that $\pi_\theta\neq\pi$. Consider
\begin{equation}
\begin{aligned}
r_\epsilon^\star
\in \operatorname*{arg\,max}_{r\in\Delta(\mathcal{V})}
&\quad
\mathbb{E}_{v\sim r}
\left[\log\frac{\pi(v)}{\pi_\theta(v)}\right] \\
\text{subject to}
&\quad D_{\mathrm{KL}}(r\|\pi_\theta)\leq\epsilon.
\end{aligned}
\end{equation}

Assume that $\epsilon>0$ is small enough that no unconstrained
maximizer of the objective is feasible. Slater's condition holds
because $\pi_\theta$ is strictly feasible and the KKT conditions
therefore apply, and the optimal trust-region multiplier
$\lambda_\epsilon^\star$ is strictly positive~\citep{boyd2004convex}.

The Lagrangian is
\begin{equation}
\begin{aligned}
\mathcal{J}_{\mathrm{TR}}(r,\lambda,\nu)
&=
\sum_{v\in\mathcal{V}}
r(v)\log\frac{\pi(v)}{\pi_\theta(v)} \\
&\quad
-\lambda\bigl(D_{\mathrm{KL}}(r\|\pi_\theta)-\epsilon\bigr) \\ 
&\quad\quad +\nu\left(\sum_{v\in\mathcal{V}}r(v)-1\right).
\end{aligned}
\end{equation}
For $\lambda_\epsilon^\star>0$, its maximizer has full support,
and stationarity gives
\begin{equation}
0
=
\log\frac{\pi(v)}{\pi_\theta(v)}
-\lambda_\epsilon^\star
\left(\log\frac{r_\epsilon^\star(v)}{\pi_\theta(v)}+1\right)
+\nu_\epsilon^\star.
\end{equation}
Rearranging and normalizing yields
\begin{equation}
r_\epsilon^\star(v)
=
\frac{
\pi_\theta(v)^{1-1/\lambda_\epsilon^\star}
\pi(v)^{1/\lambda_\epsilon^\star}
}{
\sum_{u\in\mathcal{V}}
\pi_\theta(u)^{1-1/\lambda_\epsilon^\star}
\pi(u)^{1/\lambda_\epsilon^\star}
}.
\end{equation}
Thus, setting $\alpha=1/\lambda_\epsilon^\star$ gives
\begin{equation}
r_\epsilon^\star(v)
=
\frac{\pi_\theta(v)^{1-\alpha}\pi(v)^\alpha}
{Z_{\theta,\alpha}(\alpha)}
=
\pi_{\theta,\alpha}^{\mathrm{APSD}}(v).
\end{equation}

This identifies APSD with the trust-region solution, with $\alpha$
acting as the inverse optimal multiplier. We now parameterize
this solution family by $\alpha$, rather than by the prescribed
radius $\epsilon$. Since $\lambda_\epsilon^\star>0$,
complementary slackness gives
\begin{equation}
D_{\mathrm{KL}}(r_\epsilon^\star\|\pi_\theta)=\epsilon.
\end{equation}
Accordingly, the radius associated with a chosen $\alpha$ is
\begin{equation}
\epsilon_\theta(\alpha)
:=
D_{\mathrm{KL}}
\bigl(\pi_{\theta,\alpha}^{\mathrm{APSD}}\|\pi_\theta\bigr).
\end{equation}

To characterize this induced radius, define
\begin{equation}
A_\theta(v)=\log\frac{\pi(v)}{\pi_\theta(v)},
\qquad
\psi_\theta(\alpha)=\log Z_{\theta,\alpha}.
\end{equation}
Using
\begin{equation}
\pi_{\theta,\alpha}^{\mathrm{APSD}}(v)
=
\pi_\theta(v)
\exp\bigl(\alpha A_\theta(v)-\psi_\theta(\alpha)\bigr),
\end{equation}
we obtain
\begin{equation}
\epsilon_\theta(\alpha)
=
\alpha\psi_\theta'(\alpha)-\psi_\theta(\alpha).
\end{equation}
Therefore,
\begin{equation}
\begin{aligned}
\frac{d\epsilon_\theta(\alpha)}{d\alpha}
&=
\alpha\psi_\theta''(\alpha) \\
&=
\alpha
\operatorname{Var}_{v\sim\pi_{\theta,\alpha}^{\mathrm{APSD}}}
\left[\log\frac{\pi(v)}{\pi_\theta(v)}\right]
>0,
\end{aligned}
\end{equation}
where strict positivity follows from $\alpha>0$, full support,
and $\pi_\theta\neq\pi$. Hence, increasing $\alpha$ strictly
increases the induced student-centered trust-region radius.\hfill $\square$

\subsection{Proof of Proposition 3}
\label{app:gradient}

\paragraph{Setup.}
Fix a decoding context and a coefficient $\alpha>0$. Assume that $\pi_\theta$ and $\pi$ are strictly positive distributions over the finite vocabulary $V$, and that $\pi_\theta$ is differentiable in $\theta$. The self-teacher $\pi$, the sampled context, and $\alpha$ are held fixed throughout differentiation. All gradients below are computed from the exact vocabulary-level KL divergence.

\paragraph{Gradient under full-target stop-gradient.}
Let $\pi^{\mathrm{APSD}}:=\operatorname{SG}[\pi_{\theta,\alpha}^{\mathrm{APSD}}]$ denote the detached target, which is held fixed during differentiation, so that $\nabla_\theta\pi^{\mathrm{APSD}}(v)=0$. With $\alpha$ and the self-teacher $\pi$ also held fixed, the detached-target objective is
\begin{equation}
\begin{aligned}
\mathcal{L}_{\mathrm{APSD}}^{\mathrm{SG}}
&= D_{\mathrm{KL}}\!\left(\pi_\theta\middle\|\pi^{\mathrm{APSD}}\right)
\\
&= \sum_{v\in V}\pi_\theta(v)\log\frac{\pi_\theta(v)}{\pi^{\mathrm{APSD}}(v)}.
\end{aligned}
\end{equation}

Differentiating with respect to $\theta$ while keeping $\pi^{\mathrm{APSD}}$ fixed gives
\begin{equation}
\begin{aligned}
\nabla_\theta\mathcal{L}_{\mathrm{APSD}}^{\mathrm{SG}}
&= \sum_{v\in V}\nabla_\theta\pi_\theta(v)\log\frac{\pi_\theta(v)}{\pi^{\mathrm{APSD}}(v)}
\\
&\quad + \sum_{v\in V}\pi_\theta(v)\nabla_\theta\log\pi_\theta(v)
\\
&= \sum_{v\in V}\nabla_\theta\pi_\theta(v)\left[\log\frac{\pi_\theta(v)}{\pi^{\mathrm{APSD}}(v)}+1\right].
\end{aligned}
\end{equation}

At the current parameter value, $\pi^{\mathrm{APSD}}$ has the same numerical values as $\pi_{\theta, \alpha}^{\mathrm{APSD}}$. We therefore substitute the APSD expression into the already-computed gradient using
\begin{equation}
\begin{aligned}
\log\frac{\pi_\theta(v)}{\pi^{\mathrm{APSD}}(v)}
&= \log\frac{Z_{\theta,\alpha}\pi_\theta(v)}{\pi_\theta(v)^{1-\alpha}\pi(v)^\alpha}
\\
&= \alpha\log\frac{\pi_\theta(v)}{\pi(v)}+\log Z_{\theta,\alpha}.
\end{aligned}
\end{equation}

Consequently,
\begin{equation}
\label{eq:app_detached_gradient}
\begin{aligned}
\nabla_\theta\mathcal{L}_{\mathrm{APSD}}^{\mathrm{SG}}
&= \alpha\sum_{v\in V}\nabla_\theta\pi_\theta(v)\log\frac{\pi_\theta(v)}{\pi(v)}
\\
&\quad +(\log Z_{\theta,\alpha}+1)\sum_{v\in V}\nabla_\theta\pi_\theta(v)
\\
&= \alpha\nabla_\theta D_{\mathrm{KL}}\!\left(\pi_\theta\middle\|\pi\right)
\\
&= \alpha\nabla_\theta\mathcal{L}_{\mathrm{SDPO}},
\end{aligned}
\end{equation}
where the terms independent of $v$ vanish because $\sum_{v\in V}\nabla_\theta\pi_\theta(v)=\nabla_\theta\sum_{v\in V}\pi_\theta(v)=\nabla_\theta 1=0$. Thus, full-target detachment reduces the single-step APSD gradient to an $\alpha$-scaled SDPO gradient.

\paragraph{Identifying the discarded R\'enyi contribution.}
With $\pi$ and $\alpha$ fixed, we choose the gradient-equivalent scalar potential associated with Eq.~\eqref{eq:app_detached_gradient} as
\begin{equation}
\begin{aligned}
\widetilde{\mathcal{L}}_{\mathrm{APSD}}^{\mathrm{SG}}(\theta)
&:= \alpha\mathcal{L}_{\mathrm{SDPO}}(\theta).
\end{aligned}
\end{equation}
This potential reproduces the detached-target gradient but should not be confused with the forward value of $\mathcal{L}_{\mathrm{APSD}}^{\mathrm{SG}}$, since stop-gradient preserves numerical values while changing the derivatives used for optimization.

By Proposition 1, subtracting this potential from the original objective gives
\begin{equation}
\begin{aligned}
\mathcal{L}_{\mathrm{APSD}}
-\widetilde{\mathcal{L}}_{\mathrm{APSD}}^{\mathrm{SG}}
&= \log Z_{\theta,\alpha}
\\
&= -(1-\alpha)D_\alpha(\pi\|\pi_\theta),
\end{aligned}
\end{equation}
where the second equality holds for $\alpha\neq1$. Differentiating identifies the exact contribution discarded by full-target stop-gradient:
\begin{equation}
\label{eq:app_gradient_coupling}
\begin{aligned}
\nabla_\theta\mathcal{L}_{\mathrm{APSD}}
-\nabla_\theta\mathcal{L}_{\mathrm{APSD}}^{\mathrm{SG}}
&= \nabla_\theta\log Z_{\theta,\alpha}
\\
&= -(1-\alpha)\nabla_\theta D_\alpha(\pi\|\pi_\theta).
\end{aligned}
\end{equation}

Direct differentiation of the normalizer gives
\begin{equation}
\begin{aligned}
\nabla_\theta\log Z_{\theta,\alpha}
&= \frac{1}{Z_{\theta,\alpha}}\nabla_\theta
\sum_{v\in V}\pi_\theta(v)^{1-\alpha}\pi(v)^\alpha
\\
&= (1-\alpha)\sum_{v\in V}
\pi_{\theta, \alpha}^{\mathrm{APSD}}(v)\nabla_\theta\log\pi_\theta(v).
\end{aligned}
\end{equation}
Consequently, the exact APSD gradient is
\begin{equation}
\begin{aligned}
\nabla_\theta\mathcal{L}_{\mathrm{APSD}}
&= \alpha\nabla_\theta\mathcal{L}_{\mathrm{SDPO}}
\\
&\quad +(1-\alpha)\sum_{v\in V}
\pi_{\theta, \alpha}^{\mathrm{APSD}}(v)\nabla_\theta\log\pi_\theta(v).
\end{aligned}
\end{equation}
The additional term need not be collinear with the SDPO gradient. Retaining it can therefore change the gradient direction rather than merely rescaling the detached-target update.

\paragraph{Behavior induced by the asymmetric coupling.}
To characterize this contribution, let $z(v)$ denote the student logits, with $\pi_\theta(v)=\exp(z(v))/\sum_{u\in V}\exp(z(u))$. Holding the self-teacher and $\alpha$ fixed, the softmax identity $\partial\log\pi_\theta(u)/\partial z(v)=\delta_{uv}-\pi_\theta(v)$ yields
\begin{equation}
\label{eq:app_coupling_logit}
\begin{aligned}
\frac{\partial\log Z_{\theta,\alpha}}{\partial z(v)}
&= (1-\alpha)\sum_{u\in V}
\pi_{\theta, \alpha}^{\mathrm{APSD}}(u)
\left[\delta_{uv}-\pi_\theta(v)\right]
\\
&= (1-\alpha)
\left[\pi_{\theta, \alpha}^{\mathrm{APSD}}(v)-\pi_\theta(v)\right],
\end{aligned}
\end{equation}
where $\delta_{uv}$ is the Kronecker delta. Thus, the additional descent direction in logit space is
\begin{equation}
\label{eq:app_coupling_descent}
\begin{aligned}
-\frac{\partial\log Z_{\theta,\alpha}}{\partial z(v)}
&= (\alpha-1)
\left[\pi_{\theta, \alpha}^{\mathrm{APSD}}(v)-\pi_\theta(v)\right].
\end{aligned}
\end{equation}

For correct rollouts, $0<\alpha<1$, so this correction decreases logits where the interpolated target $\pi_{\theta, \alpha_\text{correct}}^{\mathrm{APSD}}$ assigns more probability than the student and increases logits where it assigns less. It therefore counteracts target-directed shifts, providing a conservative correction consistent with preserving already successful student preferences. This correction complements, rather than replaces, the SDPO term in the full gradient.

For incorrect rollouts, $\alpha>1$, and the correction reverses direction: it increases logits where the extrapolated target $\pi_{\theta, \alpha_\text{incorrect}}^{\mathrm{APSD}}$ exceeds the student and decreases logits where the target falls below the student. The relevant discrepancy is determined by
\begin{equation}
\label{eq:app_coupling_ratio}
\begin{aligned}
\frac{\pi_{\theta, \alpha}^{\mathrm{APSD}}(v)}{\pi_\theta(v)}
&= \frac{1}{Z_{\theta,\alpha}}
\left(\frac{\pi(v)}{\pi_\theta(v)}\right)^\alpha.
\end{aligned}
\end{equation}
In particular, $\pi_{\theta, \alpha}^{\mathrm{APSD}}(v)>\pi_\theta(v)$ precisely when $\pi(v)/\pi_\theta(v)>Z_{\theta,\alpha}^{1/\alpha}$. The additional correction therefore favors alternatives with sufficiently high teacher-to-student probability ratios, rather than tokens with high teacher probability alone. A token jointly assigned high probability by the student and self-teacher is not automatically reinforced: its logit receives a negative correction whenever its ratio falls below this threshold.

The asymmetric coefficient thus switches the coupling from counteracting movement toward the interpolated target on correct rollouts to reinforcing movement toward the extrapolated target on incorrect rollouts. This provides a mechanism for conservative feedback integration after success and stronger correction after failure, provided that the self-teacher identifies useful alternatives. These statements characterize the additional logit-gradient component, not a guarantee of improved task correctness or of every token's probability change under the complete parameter update.

When $\alpha=1$, or when $\pi_\theta=\pi$, the coupling contribution vanishes. Finally, the scaled-SDPO identity holds per decoding step and for a rollout with a common $\alpha$; across rollouts with different coefficients, full-target stop-gradient yields a weighted sum of SDPO gradients rather than a single scalar multiple of the batch gradient.

\paragraph{Differentiability of the APSD target.}
The trust-region and generalized barycentric formulations both
recover the same explicit target,
\begin{equation}
\pi_{\theta,\alpha}^{\mathrm{APSD}}(v)
=
\frac{\pi_\theta(v)^{1-\alpha}\pi(v)^\alpha}
{Z_{\theta,\alpha}(\alpha)}.
\end{equation}
For fixed $\alpha$ and self-teacher distribution $\pi$, this map
is differentiable with respect to the strictly positive student
probabilities, with coordinate derivatives
\begin{equation}
\frac{\partial\pi_{\theta,\alpha}^{\mathrm{APSD}}(v)}
{\partial\pi_\theta(u)}
=
(1-\alpha)
\frac{\pi_{\theta,\alpha}^{\mathrm{APSD}}(v)}{\pi_\theta(u)}
\left[
\delta_{uv}-\pi_{\theta,\alpha}^{\mathrm{APSD}}(u)
\right],
\end{equation}
where $\delta_{uv}$ is the Kronecker delta.
Thus, characterizing the target as the solution to an auxiliary
optimization does not eliminate its dependence on the student.
Full-target stop-gradient explicitly suppresses this dependence
by setting the target Jacobian to zero during differentiation.
Throughout, $\alpha$ is held fixed, while
$\epsilon_\theta(\alpha)$ is induced by the target and may vary
with $\theta$; the radius is not held fixed as the student changes.

\paragraph{Why the envelope theorem does not justify detachment.}
The trust-region and barycentric problems are auxiliary optimizations used to construct $\pi_{\theta, \alpha}^{\mathrm{APSD}}$, whereas APSD minimizes the distinct outer objective $D_{\mathrm{KL}}(\pi_\theta\|\pi_{\theta, \alpha}^{\mathrm{APSD}})$. The envelope theorem characterizes derivatives of the auxiliary problems' optimal values; it does not justify omitting the optimizer dependence when that optimizer is reused in a different outer loss. In particular, their stationarity conditions do not cancel the target-side contribution $\nabla_\theta\log Z_{\theta,\alpha}$ identified in Eq.~\eqref{eq:app_gradient_coupling}. Full-target stop-gradient therefore leaves the forward trust-region and barycentric target unchanged, but removes its student-dependent contribution from the APSD optimization dynamics. \hfill $\square$

\section{Experimental Details}
\label{app:experimental-details}

\subsection{RLVR Training Details}
\label{app:rlvr-details}

We follow the RLVR training setup of \citet{hubotter2026reinforcement}
and summarize our training and evaluation configurations in
Table~\ref{tab:rlvr-hyperparameters}. Each training run uses four NVIDIA H200 GPUs with a wall-clock time budget of five hours.

We use a maximum prompt length of 2,048 tokens and a maximum response
length of 8,192 tokens.
Each training batch contains 32 prompts with eight rollouts per prompt.
For evaluation, we sample 16 responses per example using a temperature of
0.6 and top-$p$ of 0.95.
All methods are optimized with AdamW, weight decay of 0.01, and gradient
clipping at 1.0.

For off-policy GRPO, we use a learning rate of $1\times10^{-6}$,
whereas the remaining methods use $1\times10^{-5}$.
Following the corresponding training protocols, off-policy GRPO performs
multiple mini-batch updates per generation batch, while on-policy methods
perform a single policy update using the newly generated rollouts.
Additional method-specific configurations are listed in
Table~\ref{tab:rlvr-hyperparameters}.

\begin{table*}[t]
\centering
\footnotesize
\setlength{\tabcolsep}{6pt}
\renewcommand{\arraystretch}{1.05}

\begin{tabular}{lll}
\toprule

\textbf{Category} & \textbf{Parameter} & \textbf{Value} \\
\midrule

\multicolumn{3}{l}{\textit{Common (shared by all methods)}} \\
\midrule

\multirow{2}{*}{Data}
& Max.\ prompt length & 2048 \\
& Max.\ response length & 8192 \\
\midrule

\multirow{2}{*}{Batching}
& Question batch size & 32 \\
& Number of rollouts & 8 \\
\midrule

\multirow{2}{*}{Rollout}
& Inference engine & vLLM \\
& Temperature & 1.0 \\
\midrule

\multirow{3}{*}{Validation}
& Number of rollouts & 16 \\
& Temperature & 0.6 \\
& Top-$p$ & 0.95 \\
\midrule

\multirow{4}{*}{Training}
& Optimizer & AdamW \\
& Warmup steps & 10 \\
& Weight decay & 0.01 \\
& Gradient clip norm & 1.0 \\
\midrule

\multirow{2}{*}{Learning rate}
& GRPO (off-policy) & $1 \times 10^{-6}$ \\
& GRPO / SDPO / SRPO / RLSD / APSD & $1 \times 10^{-5}$ \\
\midrule

\multicolumn{3}{l}{\textit{GRPO / SRPO / RLSD}} \\
\midrule

\multirow{3}{*}{GRPO objective}
& $\epsilon$-high (asymmetric clip) & 0.28 \\
& KL coefficient & 0.0 \\
& Rollout IS clip ($\rho$) & 2 \\
\midrule



\multicolumn{3}{l}{\textit{APSD}} \\
\midrule
Asymmetric PoE
& $\alpha_{\mathrm{corr}}$ / $\alpha_{\mathrm{incorr}}$
& $0.9$ / $1.1$ \\

\bottomrule
\end{tabular}

\caption{
Training and evaluation hyperparameters for the five-hour RLVR experiments
with Olmo3-7B-Instruct and Qwen3-8B.
Shared configurations are listed first, followed by method-specific settings.
}
\label{tab:rlvr-hyperparameters}
\end{table*}

\subsection{RLRF Training Details}
\label{app:rlrf-details}

We follow the RLRF training setup of \citet{hubotter2026reinforcement}
and summarize the training and evaluation configurations in
Table~\ref{tab:rlrf-hyperparameters}.
All methods are trained for 80 optimization steps with 32 prompts per
batch and eight rollouts per prompt.
Training rollouts are generated with a temperature of 1.0, while
evaluation uses a temperature of 0.6 and top-$p$ of 0.95.
We use maximum prompt and response lengths of 2,048 and 8,192 tokens,
respectively.

All methods are optimized with AdamW using a learning rate of
$1\times10^{-6}$, weight decay of 0.01, and gradient clipping at 1.0.
The distribution-mixing coefficients and other method-specific
hyperparameters are reported in Table~\ref{tab:rlrf-hyperparameters}.

\begin{table*}[t]
\centering
\footnotesize
\setlength{\tabcolsep}{6pt}
\renewcommand{\arraystretch}{1.08}

\begin{tabular}{lll}
\toprule
\textbf{Category} & \textbf{Parameter} & \textbf{Value} \\
\midrule

\multicolumn{3}{l}{\textit{Common (shared by all methods)}} \\
\midrule

\multirow{2}{*}{Data}
& Max.\ prompt length & 2048 \\
& Max.\ response length & 8192 \\
\midrule

\multirow{2}{*}{Batching}
& Question batch size & 32 \\
& Training rollouts per question & 8 \\
\midrule

\multirow{2}{*}{Rollout}
& Inference engine & vLLM \\
& Temperature & 1.0 \\
\midrule

\multirow{4}{*}{Validation}
& Evaluation interval & 5 steps \\
& Rollouts per question & 4 \\
& Temperature & 0.6 \\
& Top-$p$ & 0.95 \\
\midrule

\multirow{5}{*}{Training}
& Training steps & 80 \\
& Optimizer & AdamW \\
& Learning rate & $1 \times 10^{-6}$ \\
& Weight decay & 0.01 \\
& Gradient clip norm & 1.0 \\
\midrule

\multirow{2}{*}{Warmup steps}
& GRPO / RLSD & 10 \\
& All other methods & 0 \\
\midrule

\multicolumn{3}{l}{\textit{GRPO / SRPO / RLSD}} \\
\midrule

\multirow{2}{*}{Policy objective}
& $\epsilon$-high (asymmetric clip) & 0.28 \\
& Reference-policy KL coefficient & 0.0 \\
\midrule






\multicolumn{3}{l}{\textit{Distribution-mixing parameters}} \\
\midrule

DistiLLM
& Student weight, $w$ & 0.5 \\

TAID
& $(t_{\mathrm{start}}, t_{\mathrm{end}}, \alpha, \beta)$
& $(0.4,\,1.0,\,0.0005,\,0.99)$ \\

DemoPSD
& $(\beta, w_{\max})$ & $(50,\,0.15)$ \\

APSD
& $\alpha_{\mathrm{corr}}$ / $\alpha_{\mathrm{incorr}}$
& $0.5$ / $1.5$ \\

\bottomrule
\end{tabular}

\caption{
Training and evaluation hyperparameters for the RLRF experiments on
LiveCodeBench with Qwen3-1.7B, Qwen3-4B, and Qwen3-8B.
Shared configurations are listed first, followed by method-specific
policy-optimization, distillation, and distribution-mixing settings.
}
\label{tab:rlrf-hyperparameters}
\end{table*}

\end{document}